\documentclass{article}

\PassOptionsToPackage{numbers, compress}{natbib}
\usepackage[final]{neurips_2020}

\usepackage[utf8]{inputenc} % allow utf-8 input
\usepackage[T1]{fontenc}    % use 8-bit T1 fonts
\usepackage{hyperref}       % hyperlinks
\usepackage{url}            % simple URL typesetting
\usepackage{booktabs}       % professional-quality tables
\usepackage{amsfonts}       % blackboard math symbols
\usepackage{nicefrac}       % compact symbols for 1/2, etc.
\usepackage{microtype}      % microtypography
\usepackage{subfig}
\usepackage[titletoc]{appendix}
\usepackage{amsmath}
\usepackage{amsthm}
\usepackage{array}
\usepackage{color}
\usepackage{caption}
\usepackage{mathtools}
\usepackage{mathrsfs}
\usepackage{wrapfig}
\usepackage{multirow}
\usepackage{bbm}
\usepackage{bm}
\usepackage{textcomp}
\usepackage{gensymb}
\usepackage{makecell}
\usepackage{listings}
\newtheorem{lemma}{Lemma}

\newtheorem{theorem}{Theorem}

\newtheorem{theoremappendix}{Theorem*}

\DeclareMathOperator{\R}{\mathbb{R}}

\hypersetup{
	colorlinks=true,
	urlcolor=magenta,
}

\title{Efficient Learning of Generative Models via \\ Finite-Difference Score Matching}

\renewcommand\footnotemark{}
\author{
Tianyu Pang$^{*1}$, Kun Xu$^{*1}$, Chongxuan Li$^1$, Yang Song$^2$, Stefano Ermon$^2$, Jun Zhu$^{\dagger1}$ \thanks{$^*$Equal contribution. $^\dagger$  Corresponding author.}\\
  $^{1}$Dept. of Comp. Sci. \& Tech., Institute for AI, BNRist Center,\\
Tsinghua-Bosch Joint ML Center, THBI Lab, Tsinghua University\\
  $^{2}$Department of Computer Science, Stanford University \\
  \texttt{\footnotesize pty17@mails.tsinghua.edu.cn, \{kunxu.thu, chongxuanli1991\}@gmail.com}\\
  \texttt{\footnotesize \{yangsong, ermon\}@cs.stanford.edu, dcszj@mail.tsinghua.edu.cn} \\
}

\begin{document}
% \nipsfinalcopy is no longer used

\maketitle

\begin{abstract}
Several machine learning applications involve the optimization of higher-order derivatives (e.g., gradients of gradients) during training, which can be expensive with respect to memory and computation even with automatic differentiation. As a typical example in generative modeling, score matching~(SM) involves the optimization of the trace of a Hessian. To improve computing efficiency, we rewrite the SM objective and its variants in terms of directional derivatives, and present a generic strategy to efficiently approximate any-order directional derivative with finite difference~(FD). Our approximation only involves function evaluations, which can be executed in parallel, and no gradient computations. Thus, it reduces the total computational cost while also improving numerical stability. We provide two instantiations by reformulating variants of SM objectives into the FD forms. Empirically, we demonstrate that our methods produce results comparable to the gradient-based counterparts while being much more computationally efficient.
\end{abstract}

%\vspace{-0.35cm}
\section{Introduction}
%\vspace{-0.25cm}
Deep generative models have achieved impressive progress on learning data distributions, with either an explicit density function~\citep{kingma2013auto,kingma2018glow,oord2016pixel,poon2011sum} or an implicit generative process~\citep{arjovsky2017wasserstein,goodfellow2014generative,zhu2017unpaired}. Among explicit models, energy-based models (EBMs)~\citep{lecun2006tutorial,teh2003energy} define the probability density as $p_{\theta}(x)=\widetilde{p}_{\theta}(x)/Z_{\theta}$, where $\widetilde{p}_{\theta}(x)$ denotes the unnormalized probability and $Z_{\theta} = \int \widetilde{p}_{\theta}(x) dx$ is the partition function. 
EBMs allow more flexible architectures~\citep{du2019implicit,Grathwohl2020Your} with simpler compositionality~\citep{haarnoja2017reinforcement,mnih2005learning} compared to other explicit generative models~\cite{oord2016pixel,he2019lagging}, and have better stability and mode coverage in training~\citep{kumar2019maximum,kurach2018large,xu2019understanding} compared to implicit generative models~\cite{goodfellow2014generative}. Although EBMs are appealing, training them with maximum likelihood estimate (MLE), i.e., minimizing the KL divergence between data and model distributions, is challenging because of the intractable partition function~\citep{hinton2002training}.

%Existing solutions based on the MCMC methods~\citep{du2019implicit,Grathwohl2020Your,nijkamp2019anatomy,welling2011bayesian,xie2016theory} and the variational ones~\citep{kim2016deep,kuleshov2017neural,Li2020To} are either computationally expensive or suffer from poor bias-variance trade-offs in high dimensions.
%Recent advances based on the MCMC methods ~\citep{du2019implicit,Grathwohl2020Your,nijkamp2019anatomy,welling2011bayesian,xie2016theory} successfully learn EBMs parameterized by deep neural networks. On the other hand, variational methods propose tractable bounds for $Z_{\theta}$~\citep{kim2016deep,kuleshov2017neural,Li2020To}. Whereas, these solutions are either computationally expensive or suffer from poor bias-variance trade-offs in high dimension.
% \se{can cut a bit..existing solutions based on approximate inference are either computationally expensive or suffer from poor bias-variance tradeoffs in high dimensions.}
% Various efforts have been devoted to addressing the challenge. On one hand, recent advances~\citep{du2019implicit,Grathwohl2020Your,nijkamp2019anatomy,welling2011bayesian,xie2016theory} successfully learn EBMs parameterized by deep neural networks based on MCMC methods, which are computationally expensive. On the other hand, variational methods propose tractable bounds for $Z_{\theta}$~\citep{kuleshov2017neural,Li2020To}. Compared to the MCMC methods, variational methods are computationally more efficient, but the bias and variance of the bounds make it hard to scale up to high-dimensional data.

%As a promising alternative, 
Score matching (SM)~\citep{hyvarinen2005estimation} is an alternative objective that circumvents the intractable partition function by training unnormalized models with the Fisher divergence~\citep{johnson2004information}, which depends on the Hessian trace and (Stein) score function~\citep{liu2016kernelized} of the log-density function. SM eliminates the dependence of the log-likelihood on $Z_{\theta}$ by taking derivatives w.r.t. $x$, using the fact that $\nabla_{x}\log p_{\theta}(x)=\nabla_{x}\log \widetilde{p}_{\theta}(x)$. Different variants of SM have been proposed, including approximate back-propagation~\citep{kingma2010regularized}, curvature propagation~\citep{martens2012estimating}, denoising score matching (DSM)~\citep{vincent2011connection}, a bi-level formulation for latent variable models~\citep{bao-bi} and nonparametric estimators~\citep{li2017gradient,shi2018spectral,sriperumbudur2017density,sutherland2018efficient,zhou2020nonparametric}, but they may suffer from high computational cost, biased parameter estimation, large variance, or complex implementations. Sliced score matching (SSM)~\citep{song2019sliced} alleviates these problems by providing a scalable and unbiased estimator with a simple implementation. However, most of these score matching methods optimize (high-order) derivatives of the density function, e.g., the gradient of a Hessian trace w.r.t. parameters, which are several times more computationally expensive compared to a typical end-to-end propagation, even when using reverse-mode automatic differentiation~\citep{griewank2008evaluating,paszke2019pytorch}. These extra computations need to be performed in sequential order and cannot be easily accelerated by parallel computing (as discussed in Appendix {\color{red} B.1}). Besides, the induced repetitive usage of the same intermediate results could magnify the stochastic variance and lead to numerical instability~\citep{stoer2013introduction}.

%\se{not sure what last stentece means. maybe cut? doesn't seem critical to the story. or just add "and can be numerically unstable".}

\begin{wrapfigure}{r}{0.5\textwidth}
\vspace{-0.3cm}
\centering
\includegraphics[width=0.48\textwidth]{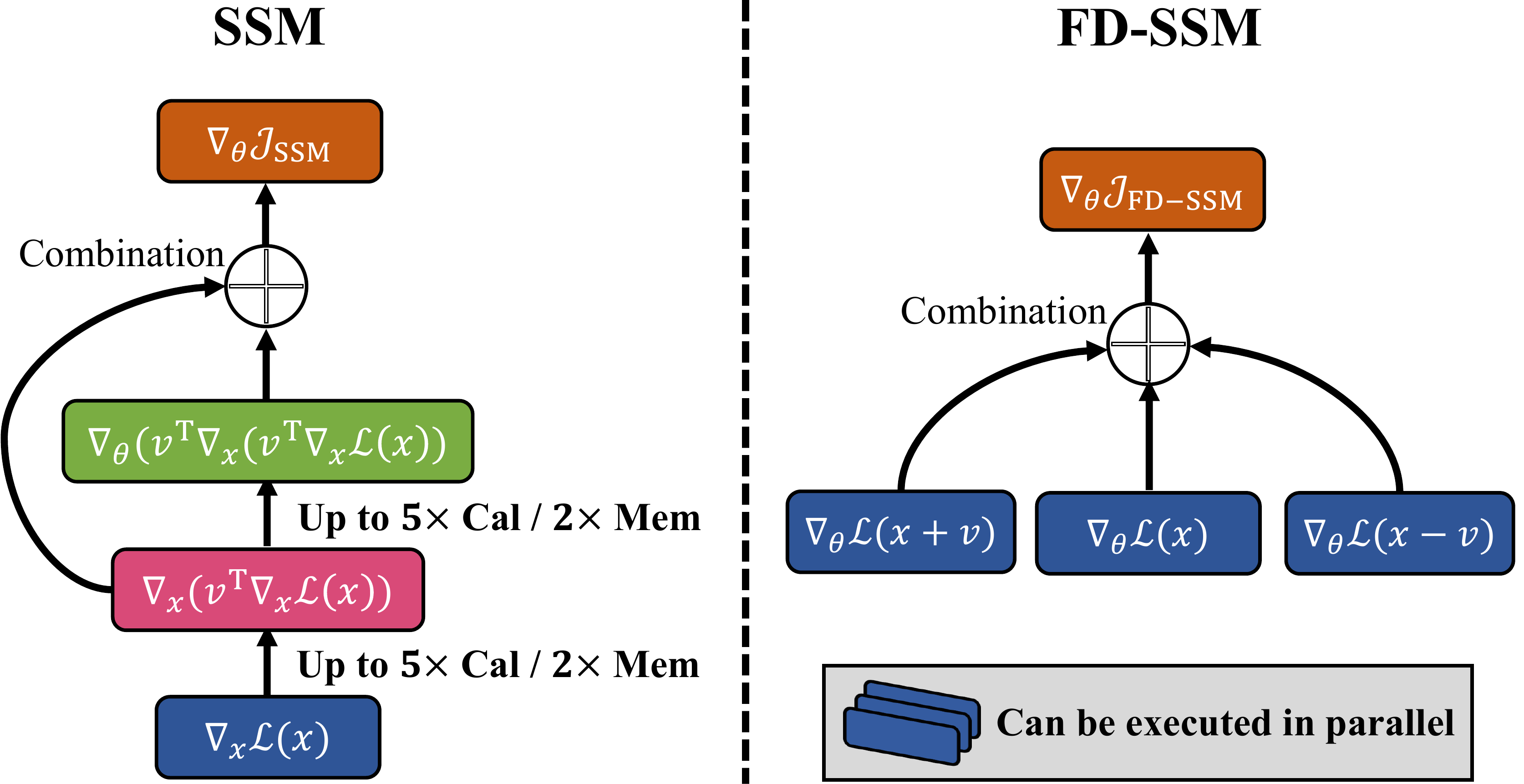}
\caption{Computing graphs of each update step. Detailed in Sec.~\ref{computational_ob} (SSM) and Sec.~\ref{FD-SSM} (FD-SSM).}
\vspace{-0.1cm}
\label{fig:approaches}
\end{wrapfigure}

To improve efficiency and stability, we first observe that existing scalable SM objectives (e.g., DSM and SSM) can be rewritten in terms of (second-order) directional derivatives. We then propose a generic finite-difference (FD) decomposition for any-order directional derivative in Sec.~\ref{FDappro}, and show an application to SM methods in Sec.~\ref{FD-SSM}, eliminating the need for optimizing on higher-order gradients. Specifically, our FD approach only requires independent (unnormalized) likelihood function evaluations, which can be efficiently and synchronously executed in parallel with a simple implementation (detailed in Sec.~\ref{computation}). This approach reduces the computational complexity of any $T$-th order directional derivative to $\mathcal{O}(T)$, and improves numerical stability because it involves a shallower computational graph. As we exemplify in Fig.~\ref{fig:approaches}, the FD reformulations decompose the inherently sequential high-order gradient computations in SSM (left panel) into simpler, independent routines (right panel). Mathematically, in Sec.~\ref{consistent} we show that even under stochastic optimization~\citep{robbins1951stochastic}, our new FD objectives are asymptotically consistent with their gradient-based counterparts under mild conditions. When the generative models are unnormalized, the intractable partition function can be eliminated by the linear combinations of log-density in the FD-form objectives. In experiments, we demonstrate the speed-up ratios of our FD reformulations with more than $2.5\times$ for SSM and $1.5\times$ for DSM on different generative models and datasets, as well as the comparable performance of the learned models.

\section{Background}
%\vspace{-0.25cm}
Explicit generative modeling aims to model the true data distribution $p_{\textup{data}}(x)$ with a parametric model $p_{\theta}(x)$, where $x\in \R^{d}$. The learning process usually minimizes some divergence between $p_{\theta}(x)$ and the (empirical) data distribution (e.g., KL-divergence minimization leads to MLE). In particular, the unnormalized generative models such as the energy-based ones~\citep{lecun2006tutorial} model the distribution as $p_{\theta}(x)=\widetilde{p}_{\theta}(x)/Z_{\theta}$, where $\widetilde{p}_{\theta}(x)$ is the unnormalized probability and $Z_{\theta} = \int \widetilde{p}_{\theta}(x) dx$ is the partition function. Computing the integral in $Z_{\theta}$ is usually intractable especially for high-dimensional data, which makes it difficult to directly learn unnormalized models with MLE~\citep{du2019implicit,kuleshov2017neural}.

% \se{define Z, seems like a key object. reiterate that computing Z is hard because it's a (high dim) integral. i would also give the MLE objective and show it depends on Z }

%\vspace{-0.15cm}
\subsection{Score matching methods}
\label{SMmethods}
%\vspace{-0.15cm}
% The partition function $Z_{\theta}$ is %existent but 
% usually intractable, which makes it difficult to directly learn unnormalized models with MLE~\citep{du2019implicit,kuleshov2017neural}. 
As an alternative to KL divergence, score matching (SM)~\citep{hyvarinen2005estimation} minimizes the Fisher divergence between $p_{\theta}(x)$ and $p_{\textup{data}}(x)$, which is equivalent to
\begin{equation}
    \mathcal{J}_{\text{SM}}(\theta)=\mathbb{E}_{p_{\textup{data}}(x)}\left[\text{tr}(\nabla_{x}^{2}\log p_{\theta}(x))+\frac{1}{2}\|\nabla_{x}\log p_{\theta}(x)\|_{2}^{2}\right]
\end{equation}
up to a constant and $\text{tr}(\cdot)$ is the matrix trace. Note that the derivatives w.r.t. $x$ eliminate the dependence on the partition function, i.e., $\nabla_{x}\log p_{\theta}(x)=\nabla_{x}\log \widetilde{p}_{\theta}(x)$, making the objective function tractable. However, the calculation of the trace of Hessian matrix is expensive, requiring the number of back-propagations proportional to the data dimension~\citep{martens2012estimating}. To circumvent this computational difficulty, two scalable variants of SM have been developed, to which we will apply our methods.
%which will be used as our concrete instantiations:
%\footnote{For notation simplicity in our derivation, we divide the DSM and SSM objectives by $d$ and ${C_{v}}$, respectively.}:

\textbf{Denoising score matching (DSM).} \citet{vincent2011connection} circumvents the Hessian trace by perturbing $x$ with a noise distribution $p_{\sigma}(\widetilde{x}|x)$ and then estimating the score of the perturbed data distribution $p_{\sigma}(\widetilde{x})=\int p_{\sigma}(\widetilde{x}|x)p_{\textup{data}}(x)dx$.
%In particular, 
When using Gaussian noise,
%$p_{\sigma}(\widetilde{x}|x)=\mathcal{N}(x,\sigma^2 I)$, we have that $\nabla_{\widetilde{x}} \log p_{\sigma}(\widetilde{x}|x)=\frac{x-\widetilde{x}}{\sigma^2}$ and 
we obtain the DSM objective as
\begin{equation}
    \mathcal{J}_{\text{DSM}}(\theta) = \frac{1}{d}\mathbb{E}_{p_{\textup{data}}(x)}\mathbb{E}_{p_{\sigma}(\widetilde{x}|x)} \left[\left\|\nabla_{\widetilde{x}} \log p_{\theta}(\widetilde{x}) + \frac{\widetilde{x}-x}{\sigma^2}\right\|_2^2\right]\text{,}
    \label{DSM}
\end{equation}
% \begin{equation}
%     \mathcal{J}_{\text{DSM}}(\theta) = \frac{1}{d}\mathbb{E}_{p_{\textup{data}}(x)}\mathbb{E}_{p_{\sigma}(\widetilde{x}|x)} \left[\|\nabla_{\widetilde{x}} \log p_{\theta}(\widetilde{x}) - \nabla_{\widetilde{x}} \log p_{\sigma}(\widetilde{x}|x)\|_2^2\right]\text{,}
%     \label{DSM}
% \end{equation}
%where the noise distribution is usually chosen as a Gaussian distribution with the covariance matrix of $\sigma I$.
The model obtained by DSM only matches the true data distribution when the noise scale $\sigma$ is small enough. However, when $\sigma\rightarrow 0$, the variance of DSM could be large or even tend to infinity~\citep{wang2020wasserstein}, requiring grid search or heuristics for choosing $\sigma$~\citep{saremi2018deep}.

\textbf{Sliced score matching (SSM).} \citet{song2019sliced} use random projections to avoid explicitly calculating the Hessian trace, so that the training objective only involves Hessian-vector products as follows:
\begin{equation}
    \mathcal{J}_{\text{SSM}}(\theta)=\frac{1}{C_{v}}\mathbb{E}_{p_{\textup{data}}(x)}\mathbb{E}_{p_{v}(v)}\left[v^{\top}\nabla^{2}_{x}\log p_{\theta}(x)v+\frac{1}{2}\left(v^{\top}\nabla_{x}\log p_{\theta}(x)\right)^{2}\right]\text{,}
    \label{eq:3}
\end{equation}
where $v\sim p_{v}(v)$ is the random direction, $\mathbb{E}_{p_{v}(v)}[vv^{\top}]\succ 0$ and $C_{v}=\mathbb{E}_{p_{v}(v)}[\|v\|_{2}^{2}]$ is a constant w.r.t. $\theta$. We divide the SSM loss by $C_{v}$ to exclude the dependence on the scale of the projection distribution $p_{v}(v)$. Here $p_{\textup{data}}(x)$ and $p_{v}(v)$ are independent. Unlike DSM, the model obtained by SSM can match the original unperturbed data distribution, but requires more expensive, high-order derivatives.

\vspace{-0.1cm}
\subsection{Computational cost of gradient-based SM methods}
\vspace{-0.1cm}
\label{computational_ob}
Although SM methods can bypass the intractable partition function $Z_{\theta}$, they have to optimize an objective function involving higher-order derivatives of the log-likelihood density. Even if reverse mode automatic differentiation is used~\citep{paszke2019pytorch}, existing SM methods like DSM and SSM can be computationally expensive during training when calculating the Hessian-vector products.

\textbf{Complexity of the Hessian-vector products.} Let $\mathcal{L}$ be any loss function, and let $\textup{Cal}(\nabla\mathcal{L})$ and $\textup{Mem}(\nabla\mathcal{L})$ denote the time and memory required to compute $\nabla\mathcal{L}$, respectively. Then if the reverse mode of automatic differentiation is used, the Hessian-vector product can be computed with up to five times more time and two times more memory compared to $\nabla\mathcal{L}$, i.e., $5\!\times\textup{Cal}(\nabla\mathcal{L})$ time and $2\!\times\textup{Mem}(\nabla\mathcal{L})$ memory~\citep{griewank1993some,griewank2008evaluating}. When we instantiate $\mathcal{L}=\log p_{\theta}(x)$, we can derive that the computations of optimizing DSM and SSM are separately dominated by the sequential operations of $\nabla_{\theta}(\|\nabla_{x}\mathcal{L}\|)$ and $\nabla_{\theta}(v^{\top}\nabla_{x}(v^{\top}\nabla_{x}\mathcal{L}))$, as illustrated in Fig.~\ref{fig:approaches} for SSM.
The operations of $\nabla_{\theta}$ and $\nabla_{x}$ require comparable computing resources, so we can conclude that compared to directly optimizing the log-likelihood, DSM requires up to $5\times$ computing time and $2\times$ memory, while SSM requires up to $25\times$ computing time and $4\times$ memory~\citep{griewank1993some}. For higher-order derivatives, we empirically observe that the computing time and memory usage grow exponentially w.r.t. the order of derivatives, i.e., the times of executing the operator $v^{\top}\nabla$, as detailed in Sec.~\ref{computation}.

\vspace{-0.2cm}
\section{Approximating directional derivatives via finite difference}
\vspace{-0.2cm}
\label{FDappro}
In this section, we first rewrite the most expensive terms in the SM objectives in terms of directional derivatives, then we provide generic and efficient formulas to approximate any $T$-th order directional derivative using finite difference (FD). The proposed FD approximations decompose the sequential and dependent computations of high-order derivatives into independent and parallelizable computing routines, reducing the computational complexity to $\mathcal{O}(T)$ and improving numerical stability.
%, and can easily be parallelized in a synchronous manner. 

% The operator $v^{\top}\nabla_{x}$ first explicitly executes the gradient operator $\nabla_{x}$ along $d$ coordinate directions and then projects onto the vector $v$, which requires $d$ times computation compared to $\frac{\partial}{\partial v}$. However, it is non-trivial to analytically compute $\frac{\partial}{\partial v}$, while the high-order directional derivatives still requires sequentially order-by-order executions. 

\vspace{-0.1cm}
\subsection{Rewriting SM objectives in directional derivatives}
\vspace{-0.1cm}
Note that the objectives of SM, DSM, and SSM described in Sec.~\ref{SMmethods} can all be abstracted in terms of $v^{\top}\nabla_{x}\mathcal{L}_{\theta}(x)$ and $v^{\top}\nabla_{x}^{2}\mathcal{L}_{\theta}(x)v$. Specifically, as to SM or DSM, $v$ is the basis vector $\bm{e}_{i}$ along the $i$-th coordinate to constitute the squared norm term $\|\nabla_{x}\mathcal{L}_{\theta}(x)\|_{2}^{2}=\sum_{i=1}^{d}(\bm{e}_{i}^{\top}\nabla_{x}\mathcal{L}_{\theta}(x))^{2}$ or the Hessian trace term $\textup{tr}(\nabla_{x}^2\mathcal{L}_{\theta}(x))=\sum_{i=1}^{d}\bm{e}_{i}^{\top}\nabla_{x}^{2}\mathcal{L}_{\theta}(x)\bm{e}_{i}$. As to SSM, $v$ denotes the random direction.

We regard the gradient operator $\nabla_{x}$ as a $d$-dimensional vector $\nabla_{x}=(\frac{\partial}{\partial x_{1}},\cdots,\frac{\partial}{\partial x_{d}})$, and  $v^{\top}\nabla_{x}$ is an operator that first executes $\nabla_{x}$ and then projects onto the vector $v$. For notation simplicity, we denote $\|v\|_{2}=\epsilon$ and rewrite the above terms as (higher-order) directional derivatives as follows:
% We regard the gradient operator $\nabla_{x}$ as a $d$-dimensional vector as $\nabla_{x}=(\frac{\partial}{\partial x_{1}},\cdots,\frac{\partial}{\partial x_{d}})$, and  $v^{\top}\nabla_{x}$ is an operator that first explicitly executes the gradient operator $\nabla_{x}$ along $d$ coordinate directions and then projects onto the vector $v$. For notation simplicity, we denote $\|v\|_{2}=\epsilon$ and rewrite the above terms as (higher-order) directional derivatives as follows:
\begin{equation}
    \! v^{\top}\nabla_{x}=\epsilon\frac{\partial}{\partial v}\text{; }\;\;\; v^{\top}\nabla_{x}\mathcal{L}_{\theta}(x)\!=\!\epsilon\frac{\partial}{\partial v}\mathcal{L}_{\theta}(x)\text{; }\;\;\; v^{\top}\nabla_{x}^{2}\mathcal{L}_{\theta}(x)v\!=\!(v^{\top}\nabla_{x})^{2}\mathcal{L}_{\theta}(x)\!=\!\epsilon^{2}\frac{\partial^{2}}{\partial v^{2}}\mathcal{L}_{\theta}(x)\text{.}\!
    \label{directional}
\end{equation}
Here $\frac{\partial}{\partial v}$ is the directional derivative along $v$, and $\left(v^{\top}\nabla_{x}\right)^{2}$ means executing 
%the operation
$v^{\top}\nabla_{x}$ twice.

\vspace{-0.1cm}
\subsection{FD decomposition for directional derivatives}
\vspace{-0.1cm}
\label{FDdecom}
%In light of the computational challenges of SM methods described in Sec.~\ref{computational_ob}, 
We propose to adopt the FD approach, a popular tool in numerical analysis to approximate differential operations~\citep{stoer2013introduction}, to efficiently estimate the terms in Eq.~(\ref{directional}). Taking the first-order case as an example, the key idea is that we can approximate $\frac{\partial}{\partial v}\mathcal{L}_{\theta}(x)\!=\!\frac{1}{2\epsilon}(\mathcal{L}_{\theta}(x\!+\!v)\!-\!\mathcal{L}_{\theta}(x\!-\!v))\!+\!o(\epsilon)$, where the right-hand side does not involve derivatives, just function evaluations. 
In FD, $\|v\|_{2}=\epsilon$ is assumed to be a small value, but this does not affect the optimization of SM objectives. For instance, the SSM objective in Eq.~(\ref{eq:3}) can be adaptively rescaled by $C_{v}$ (generally explained in Appendix {\color{red} B.2}).

%For example, we can recover the SSM objective in Eq.~(\ref{eq:3}) by dividing the factor $C_{v}$.

%\se{explain this homogeneity business/rescaling with an extra sentence}

%Generically, as to the $T$-th order directional derivative, 
In general, to estimate the $T$-th order directional derivative of $\mathcal{L}_\theta$, which is assumed to be $T$ times differentiable, we first apply the multivariate Taylor's expansion with Peano's remainder~\citep{konigsberger2004analysis} as 
\begin{equation}
    \mathcal{L}_{\theta}(x+\gamma v)=\sum_{t=0}^{T}\frac{\gamma^{t}}{t!}\left(v^{\top}\nabla_{x}\right)^{t}\mathcal{L}_{\theta}(x)+o(\epsilon^{T})=\sum_{t=0}^{T}\gamma^{t}\left(\frac{\epsilon^t}{t!}\frac{\partial^{t}}{\partial v^{t}}\mathcal{L}_{\theta}(x)\right)+o(\epsilon^{T})\text{,}\!
    \label{expansion}
\end{equation}
where $\gamma\in\R$ is a certain coefficient. Then, we take a linear combination of the Taylor expansion in Eq.~(\ref{expansion}) for different values of $\gamma$ and eliminate derivative terms of order less than $T$. Formally, $T\!+\!1$ different $\gamma$s are sufficient to construct a valid FD approximation (all the proofs are in Appendix {\color{red} A}).\footnote{Similar conclusions as
in Lemma~\ref{theo_lemma} and Theorem~\ref{theo} were previously found in the Chapter {\color{red} 6.5} of \citet{isaacson2012analysis} under the univariate case, while we generalize them to the multivariate case.}
%we can exploit a set of $K$ different coefficients $\{\gamma_{i}\}_{i=1}^{K}$ and leverage the linear combination of $\mathcal{L}_{\theta}(x+\gamma_{i} v)$ to eliminate the $t$-th term in Eq.~(\ref{expansion}) for any $t<T$. The following lemma states that we need at most $K\!=\!T\!+\!1$ to construct a valid FD approximation (all the proofs are provided in Appendix {\color{red} A}):
\begin{lemma}
\label{theo_lemma}
(Existence of $o(1)$ estimator) If $\mathcal{L}_{\theta}(x)$ is $T$-times-differentiable at $x$, then given any set of $T+1$ different real values $\{\gamma_{i}\}_{i=1}^{T+1}$, there exist corresponding coefficients $\{\beta_{i}\}_{i=1}^{T+1}$, such that 
\begin{equation}
    \frac{\partial^{T}}{\partial v^{T}}\mathcal{L}_{\theta}(x)=\frac{T!}{\epsilon^{T}}\sum_{i=1}^{T+1}\beta_{i}\mathcal{L}_{\theta}(x+\gamma_{i}v)+o(1)\text{.}
\end{equation}
% \vspace{-0.1cm}
\end{lemma}
%\se{there is a high chance a result like this is in some textbook. if so we shouldn't claim it's new, and we should reference some books.}
% \begin{lemma}
% \label{theo_lemma}
% (Proof in Appendix {\color{red} A.1}) If $\mathcal{L}_{\theta}(x)$ is $T$-times-differentiable at $x$, then given any set of $T+1$ different real values $\{\gamma_{i}\}_{i\in[T+1]}$ and the induced Vandermonde matrix $V\in\R^{(T+1)\times (T+1)}$, where the element $V_{i,j}=\gamma_{i}^{j-1}$, we can provide the FD approximation as
% \begin{equation}
%     \frac{\partial^{T}}{\partial v^{T}}\mathcal{L}_{\theta}(x)=\frac{T!}{\epsilon^{T}}\sum_{i=1}^{T+1}\beta_{i}\mathcal{L}_{\theta}(x+\gamma_{i}v)+o(1)\text{,}
% \end{equation}
% where $\|v\|_{2}=\epsilon$, $\bm{\beta}\in\R^{T+1}$, and $V^{\top}\bm{\beta}=\bm{e_{(T+1)}}$ is the one-hot vector of the $(T\!+\!1)$-th element.
% \end{lemma}
Lemma~\ref{theo_lemma} states that it is possible to approximate the $T$-th order directional derivative as to an $o(1)$ error
with $T\!+\!1$ function evaluations. In fact, as long as $\mathcal{L}_{\theta}(x)$ is $(T\!+\!1)$-times-differentiable at $x$, we can construct a special kind of linear combination of $T\!+\!1$ function evaluations to reduce the approximation error to $o(\epsilon)$, as stated below:
\begin{theorem}
\label{theo}
(Construction of $o(\epsilon)$ estimator) If $\mathcal{L}_{\theta}(x)$ is $(T\!+\!1)$-times-differentiable at $x$, we let $K\in \mathbb{N}^{+}$ and $\{\alpha_{k}\}_{k=1}^{K}$ be any set of $K$ different positive numbers, then we have the FD decomposition
\begin{equation}
    \!\frac{\partial^{T}}{\partial v^{T}}\mathcal{L}_{\theta}(x)\!=\!o(\epsilon)\!+\!\left\{ 
\begin{aligned}
    &\!\frac{T!}{2\epsilon^{T}}\!\sum_{k\in[K]}\beta_{k}\alpha_{k}^{-2}\left[\mathcal{L}_{\theta}(x\!+\!\alpha_{k}v)\!+\!\mathcal{L}_{\theta}(x\!-\!\alpha_{k}v)\!-\!2\mathcal{L}_{\theta}(x)\right]\text{, when }T\!=\!2K\textup{;}\!\!\\
    &\!\frac{T!}{2\epsilon^{T}}\!\sum_{k\in[K]}\beta_{k}\alpha_{k}^{-1}\left[\mathcal{L}_{\theta}(x\!+\!\alpha_{k}v)\!-\!\mathcal{L}_{\theta}(x\!-\!\alpha_{k}v)\right]\text{, when }T\!=\!2K-1\textup{.}\!\!
\end{aligned}
\right.
\end{equation}
The coefficients $\bm{\beta}\in\R^{K}$ is the solution of $V^{\top}\bm{\beta}=\bm{e}_{K}$, where $V\in\R^{K\times K}$ is the Vandermonde matrix induced by $\{\alpha_{k}^{2}\}_{k=1}^{K}$, i.e., $V_{ij}=\alpha_{i}^{2j-2}$, and $\bm{e}_{K}\in\R^{K}$ is the $K$-th basis vector.
\end{theorem}
%\se{same..if known (and i suspect it is), cite a book}
%The choice of each $\alpha_{i}$ is arbitrary as long as they are different, such that the Vandermonde matrix $V$ is non-singular.
%\se{move this sentence to experiments, or comment out. breaks the flow}
% \se{these theorems are hard to parse. might be worth putting in appendix, and putting an informal statement here. like "given the value of L evaluated at xxx different points, we can get a xxx order approx to the directional derivative by taking a linear combination of ..". or something like that}
% \se{should give some intuition for these theorems and where they are going. we can approximate a (higher order) partial derivative by evaluating the function in a neighborhood of x. similar to approximating f' with f(x)-f(x+epsilon).. RHS does not involve gradients, just function evaluation}
It is easy to generalize Theorem~\ref{theo} to achieve approximation error $o(\epsilon^{N})$ for any $N\geq 1$ with $T\!+\!N$ function evaluations, and we can show that the error rate $o(\epsilon)$ is optimal when evaluating $T\!+\!1$ functions. So far we have proposed generic formulas for the FD decomposition of any-order directional derivative. As to the application to SM objectives (detailed in Sec.~\ref{FD-SSM}), we can instantiate the decomposition in Theorem~\ref{theo} with $K=1$, $\alpha_{1}=1$, and solve for $\beta_{1}=1$, which leads to
\begin{equation}
\left\{ 
\begin{aligned}
    &v^{\top}\nabla_{x}\mathcal{L}_{\theta}(x)=\epsilon\frac{\partial}{\partial v}\mathcal{L}_{\theta}(x)=\frac{1}{2}\mathcal{L}_{\theta}(x\!+\!v)-\frac{1}{2}\mathcal{L}_{\theta}(x\!-\!v)+o(\epsilon^2)\text{;}\\
    &v^{\top}\nabla^{2}_{x}\mathcal{L}_{\theta}(x)v=\epsilon^{2}\frac{\partial^{2}}{\partial v^{2}}\mathcal{L}_{\theta}(x)=\mathcal{L}_{\theta}(x\!+\!v)+\mathcal{L}_{\theta}(x\!-\!v)-2\mathcal{L}_{\theta}(x)+o(\epsilon^3)\text{.}
\end{aligned}
\right.
\label{appro}
\end{equation}
In addition to generative modeling, the decomposition in Theorem~\ref{theo} can potentially be used in other settings involving higher-order derivatives, e.g., extracting local patterns with high-order directional derivatives~\citep{yuan2014rotation}, training GANs with gradient penalty~\citep{mescheder2018training}, or optimizing the Fisher information~\citep{brunel1998mutual}. We leave these interesting explorations to future work.

\label{3.1remark}
\textbf{Remark.} When $\mathcal{L}_{\theta}(x)$ is modeled by a neural network, we can employ the average pooling layer and the non-linear activation of, e.g., Softplus~\citep{zheng2015improving} to have an infinitely differentiable model to meet the condition in Theorem~\ref{theo}. Note that Theorem~\ref{theo} promises a \emph{point-wise} approximation error $o(\epsilon)$. To validate the error rate \emph{under expectation} for training objectives, we only need to assume that $p_{\textup{data}}(x)$ and $\mathcal{L}_{\theta}(x)$ satisfy mild regularity conditions beyond the one in Theorem~\ref{theo}, which can be easily met in practice, as detailed in Appendix {\color{red} B.3}. Conceptually, these mild regularity conditions enable us to substitute the Peano's remainders with Lagrange's ones. Moreover, this substitution results in a better approximation error of $\mathcal{O}(\epsilon^2)$ for our FD decomposition, while we still use $o(\epsilon)$ for convenience.

%\se{one question i have is to what extent the homogeneity/rescaling trick is generally applicable}

\vspace{-0.11cm}
\subsection{Computational efficiency of the FD decomposition}
\vspace{-0.11cm}
\label{computation}

\begin{wrapfigure}{r}{0.5\textwidth}
\vspace{-0.3cm}
\centering
\includegraphics[width=0.5\textwidth]{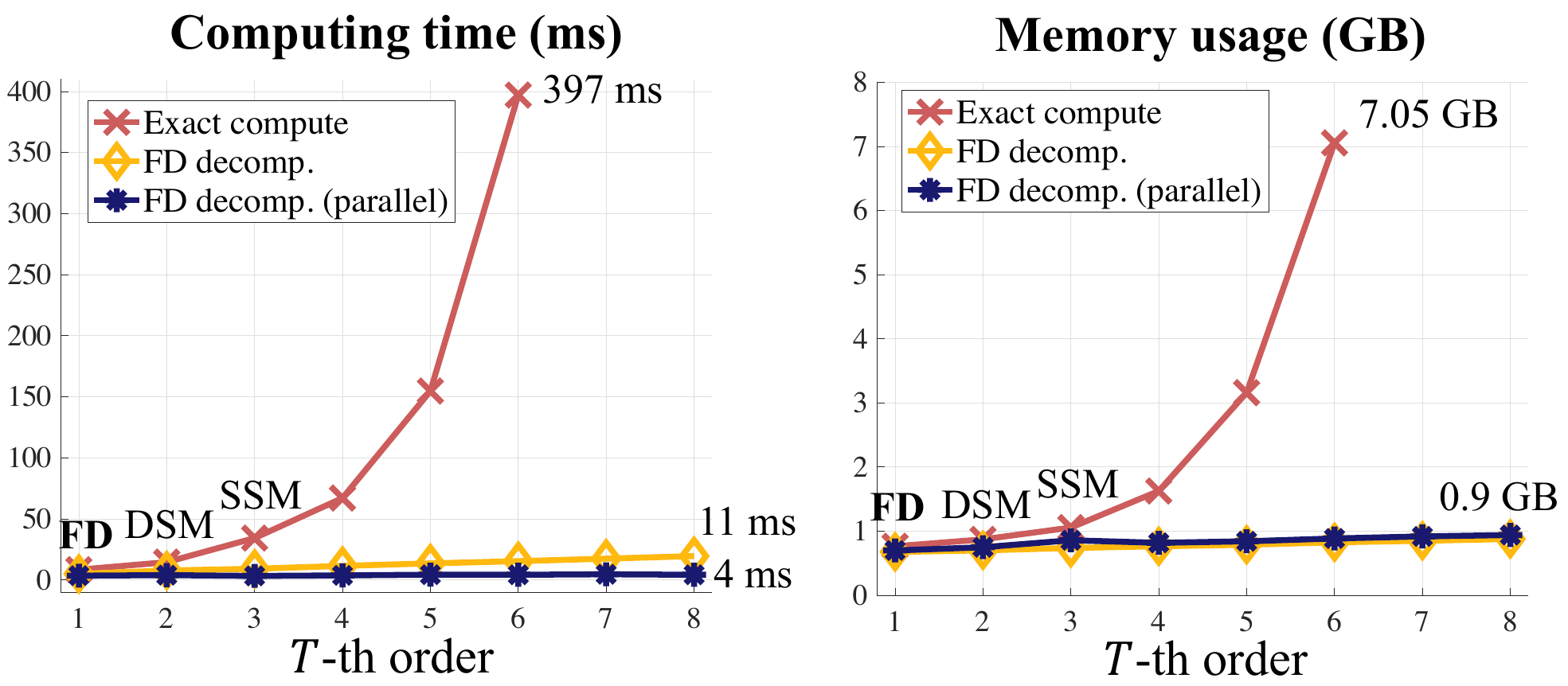}
\vspace{-0.5cm}
\caption{Computing time and memory usage for calculating the $T$-th order directional derivative.}
\label{fig:demo}
\end{wrapfigure}

Theorem~\ref{theo} provides a generic approach to approximate any $T$-th order directional derivative by decomposing the sequential and dependent order-by-order computations into independent function evaluations. This decomposition reduces the computational complexity to $\mathcal{O}(T)$, while the complexity of explicitly computing high-order derivatives usually grows exponentially w.r.t. $T$~\citep{griewank1993some}, as we verify in Fig.~\ref{fig:demo}. Furthermore, due to the mutual independence among the function terms $\mathcal{L}_{\theta}(x+\gamma_{i}v)$, they can be efficiently and synchronously executed in parallel via simple implementation (pseudo code is in Appendix {\color{red} C.1}). Since this parallelization acts on the level of operations for each data point $x$, it is compatible with data or model parallelism to further accelerate the calculations.

To empirically demonstrate the computational efficiency of our FD decomposition, we report the computing time and memory usage in Fig.~\ref{fig:demo} for calculating the $T$-th order directional derivative, i.e., $\frac{\partial^{T}}{\partial v^{T}}$ or $(v^{\top}\nabla_{x})^{T}$, either exactly or by the FD decomposition. The function $\mathcal{L}_{\theta}(x)$ is the log-density modeled by a deep EBM and trained on MNIST, while we use PyTorch~\citep{paszke2019pytorch} for automatic differentiation. As shown in the results, our FD decomposition significantly promotes efficiency in respect of both speed and memory usage, while the empirical approximation error rates are kept within $1\%$. When we parallelize the FD decomposition, the computing time is almost a constant w.r.t. the order $T$, as long as there is enough GPU memory. In our experiments in Sec.~\ref{exp}, the computational efficiency is additionally validated on the FD-reformulated SM methods.

\vspace{-0.1cm}
\section{Application to  score matching methods}
% \se{consider splitting this into a section, it's already too long and too much content}
\vspace{-0.1cm}
\label{FD-SSM}
% We apply specific instantiations of our FD formulas to DSM and SSM, thus accelerating and stabilizing the learning procedure by not explicitly optimizing on the higher-order derivatives.

Now we can instantiate $\mathcal{L}_{\theta}(x)$ in Eq.~(\ref{appro}) as the log-density function $\log p_{\theta}(x)$ to reformulate the gradient-based SM methods. For unnormalized models $p_{\theta}(x)=\widetilde{p}_{\theta}(x)/Z_{\theta}$, the decomposition in Theorem~\ref{theo} can naturally circumvent $Z_{\theta}$ by, e.g., $\log p_{\theta}(x+\alpha_{k}v)-\log p_{\theta}(x-\alpha_{k}v)=\log \widetilde{p}_{\theta}(x+\alpha_{k}v)-\log \widetilde{p}_{\theta}(x-\alpha_{k}v)$ where the partition function term cancels out, even without taking derivatives. Thus, the FD reformulations introduced in this section maintain the desirable property of their gradient-based counterparts of bypassing the intractable partition function. For simplicity, we set the random projection $v$ to be uniformly distributed as $p_{\epsilon}(v)=\mathcal{U}(\{v\in\R^{d}|\|v\|=\epsilon\})$, while our conclusions generally hold for other distributions of $v$ with bounded support sets.

%Now we can apply the FD decomposition in Eq.~(\ref{appro}) to reformulate the gradient-based SM methods, where the generic function $\mathcal{L}_{\theta}(x)$ is instantiated as the log-density function $\log p_{\theta}(x)$.

% \se{there's a lot of math below likely making this unreadable, consider moving to appendix as much as possible, keeping just a few key things. for example, i doubt people will care much about the difference between ssm and ssmvr..as long as they get the main idea }

% In the following, we set the random projection $v$ to uniformly distribute on a small hypersphere of radius $\epsilon$ by default, i.e., $p_{\epsilon}(v)=\mathcal{U}(\{v\in\R^{d}|\|v\|=\epsilon\})$, where $\epsilon$ is a small value. We assume that the model $p_{\theta}(x)$ always satisfies the differentiability condition in Theorem~\ref{theo}. Our conclusions generally hold for other distributions of $v$ with bounded support sets.

\textbf{Finite-difference SSM.} For SSM, the scale factor is $C_{v}=\epsilon^2$ in Eq.~(\ref{eq:3}). By instantiating $\mathcal{L}_{\theta}=\log p_{\theta}(x)$ in Eq.~(\ref{appro}), we propose the finite-difference SSM (\textbf{FD-SSM}) objective as
\begin{equation}
\begin{split}
  \!\!  \mathcal{J}_{\text{FD-SSM}}(\theta)=\frac{1}{\epsilon^2}\mathbb{E}_{p_{\textup{data}}(x)}\mathbb{E}_{p_{\epsilon}(v)}\Big[&\log p_{\theta}(x+v)+\log p_{\theta}(x-v)-2\log p_{\theta}(x)\\
    &+\frac{1}{8}\left(\log p_{\theta}(x+v)-\log p_{\theta}(x-v)\right)^{2}\Big]=\mathcal{J}_{\text{SSM}}(\theta)+{\color{blue}o(\epsilon)}\text{.}\!\!
\end{split}
 \label{ESM}
\end{equation}
In Fig.~\ref{fig:approaches}, we intuitively illustrate the computational graph to better highlight the difference between the gradient-based objectives and their FD reformations, taking SSM as an example.

\textbf{Finite-difference DSM.} To construct the FD instantiation for DSM, we first cast the original objective in Eq.~(\ref{DSM}) into sliced Wasserstein distance~\citep{rabin2011wasserstein} with random projection $v$ (detailed in Appendix {\color{red} B.4}).
% Since there is $\mathbb{E}_{p_{\epsilon}(v)}\left[vv^{\top}\right]=\frac{\epsilon^2I}{d}$, we can rewrite the objective of DSM with Gaussian noise distribution as
% \begin{equation}
%     \mathcal{J}_{\text{DSM}}(\theta) = \frac{1}{\epsilon^2} \mathbb{E}_{p_{\textup{data}}(x)}\mathbb{E}_{p_{\sigma}(\widetilde{x}|x)}\mathbb{E}_{p_{\epsilon}(v)} \left[\left(v^{\top}\nabla_{\widetilde{x}} \log p_{\theta}(\widetilde{x}) + \frac{v^{\top}(\widetilde{x}-x)}{\sigma^2}\right)^2\right]\text{.}
%     \label{DSM_trace}
% \end{equation}
% In this case, there is $\frac{v^{\top}(\widetilde{x}-x)}{\sigma^2}=\mathcal{O}(\epsilon)$ with high probability, thus we can approximate $v^{\top}\nabla_{\widetilde{x}} \log p_{\theta}(\widetilde{x})$ according to Eq.~(\ref{appro}) and 
Then we can propose the finite-difference DSM (\textbf{FD-DSM}) objective as
\begin{equation}
  \mathcal{J}_{\text{FD-DSM}}(\theta)\!=\! \frac{1}{4\epsilon^2}\mathbb{E}_{p_{\textup{data}}(x)}\mathbb{E}_{p_{\sigma}(\widetilde{x}|x)}\mathbb{E}_{p_{\epsilon}(v)}\!\left[ \left(\log p_{\theta}(\widetilde{x}\!+\!v)\!-\!\log p_{\theta}(\widetilde{x}\!-\!v)\!+\!\frac{2v^{\top}(\widetilde{x}\!-\!x)}{\sigma^2}\right)^2\right]\!\text{.}\!
    \label{eq:18}
\end{equation}
It is easy to verify that $\mathcal{J}_{\text{FD-DSM}}(\theta)=\mathcal{J}_{\text{DSM}}(\theta) + {\color{blue}o(\epsilon)}$, and we can generalize FD-DSM to the cases with other noise distributions of $p_{\sigma}(\widetilde{x}|x)$ using similar instantiations of Eq.~(\ref{appro}).

\textbf{Finite-difference SSMVR.} Our FD reformulation can also be used for \emph{score-based generative models}~\citep{saremi2019neural,song2019generative}, where $s_{\theta}(x):\R^d\rightarrow\R^d$ estimates $\nabla_{x}\log p_{\textup{data}}(x)$ without modeling the likelihood by $p_{\theta}(x)$. In this case, we utilize the fact that $\mathbb{E}_{p_{\epsilon}(v)}\left[vv^{\top}\right]=\frac{\epsilon^2I}{d}$ and focus on the objective of SSM with variance reduction (SSMVR)~\citep{song2019sliced}, where $\frac{1}{\epsilon^2}\mathbb{E}_{p_{\epsilon}(v)}[(v^{\top}s_{\theta}(x))^{2}]=\frac{1}{d}\|s_{\theta}(x)\|_{2}^{2}$ as
\begin{equation}
    \mathcal{J}_{\text{SSMVR}}(\theta)=\mathbb{E}_{p_{\textup{data}}(x)}\mathbb{E}_{p_{\epsilon}(v)}\left[\frac{1}{\epsilon^2}v^{\top}\nabla_{x}s_{\theta}(x)v+\frac{1}{2d}\|s_{\theta}(x)\|_{2}^{2}\right]\text{.}
\end{equation}
If $s_{\theta}(x)$ is (element-wisely) twice-differentiable at $x$, we have the expansion that $s_{\theta}(x+v)+s_{\theta}(x-v)=2s_{\theta}(x)+o(\epsilon)$ and $s_{\theta}(x+v)-s_{\theta}(x-v)=2\nabla_{x}s_{\theta}(x)v+o(\epsilon^2)$. Then we can construct the finite-difference SSMVR (\textbf{FD-SSMVR}) for the score-based models as
\begin{equation*}
        \mathcal{J}_{\text{FD-SSMVR}}(\theta)\!=\!\mathbb{E}_{p_{\textup{data}}(x)}\mathbb{E}_{p_{\epsilon}(v)}\!\left[\frac{1}{8d}\|s_{\theta}(x\!+\!v)\!+\!s_{\theta}(x\!-\!v)\|_{2}^{2}\!+\!\frac{1}{2\epsilon^{2}}\!\left(v^{\top}s_{\theta}(x\!+\!v)\!-\!v^{\top}s_{\theta}(x\!-\!v)\right)\right]\text{.}
        \label{ESM_scorenet_VR}
\end{equation*}
We can verify that $\mathcal{J}_{\text{FD-SSMVR}}(\theta)=\mathcal{J}_{\text{SSMVR}}(\theta)+{\color{blue}o(\epsilon)}$. Compared to the FD-SSM objective on the likelihood-based models, we only use two counterparts $s_{\theta}(x+v)$ and $s_{\theta}(x-v)$ in this instantiation.

\vspace{-0.2cm}
%\section{Theoretical analysis}
\section{Consistency under stochastic optimization}
\vspace{-0.2cm}
\label{consistent}
In practice, we usually apply mini-batch stochastic gradient descent (SGD)~\citep{robbins1951stochastic} to update the model parameters $\theta$. Thus beyond the expected $o(\epsilon)$ approximation error derived in Sec.~\ref{FD-SSM}, it is critical to formally verify the consistency between the FD-form objectives and their gradient-based counterparts under stochastic optimization. To this end, we establish a uniform convergence theorem for FD-SSM as an example, while similar proofs can be applied to other FD instantiations as detailed in Appendix {\color{red}B.5}. A key insight is to show that the directions of $\nabla_{\theta}\mathcal{J}_{\textup{FD-SSM}}(\theta)$ and $\nabla_{\theta}\mathcal{J}_{\textup{SSM}}(\theta)$ are sufficiently aligned under SGD, as stated in Lemma~\ref{theorem1}:

% \begin{theorem}
% \label{theorem1}
% (Proof in Appendix A.1) Under the condition that $\log p_{\theta}(x)$ is four-times-differentiable at $(x,\theta)$, we have
% \begin{equation}
%     \frac{1}{\epsilon^2}\|\nabla_{\theta}\mathcal{J}_{\textup{FD-SSM}}(\theta)-\nabla_{\theta}\mathcal{J}_{\textup{SSM}}(\theta)\|_{\infty}=o(\epsilon)\text{.}
% \end{equation}
% This means that after scaling with a factor of $1/\epsilon^2$, the gradients of the SSM and the FD-SSM objectives w.r.t. $\theta$ are consistent with $\epsilon\rightarrow 0$.
% \end{theorem}

\begin{lemma}
\label{theorem1}
(Uniform guarantee) Let $\mathcal{S}$ be the parameter space of $\theta$, $B$ be a bounded set in the space of $\R^{d}\times\mathcal{S}$, and $B_{\epsilon_{0}}$ be the $\epsilon_{0}$-neighbourhood of $B$ for certain $\epsilon_{0}>0$. Then under the condition that $\log p_{\theta}(x)$ is four times continuously differentiable w.r.t. $(x,\theta)$ and $\|\nabla_{\theta}\mathcal{J}_{\textup{SSM}}(x,v;\theta)\|_{2}>0$ in the closure of $B_{\epsilon_{0}}$, we have $\forall \eta>0$, $\exists \xi>0$, such that
\begin{equation}
    \angle\left(\nabla_{\theta}\mathcal{J}_{\textup{FD-SSM}}(x,v;\theta), \nabla_{\theta}\mathcal{J}_{\textup{SSM}}(x,v;\theta)\right)<\eta
\end{equation}
uniformly holds for $\forall(x,\theta)\in B, v\in\R^{d}, \|v\|_{2}=\epsilon<\min(\xi,\epsilon_{0})$. Here $\angle(\cdot,\cdot)$ denotes the angle between two vectors. The arguments $x,v$ in the objectives indicate the losses at that point.
\end{lemma}

Note that during the training process, we do not need to define a specific bounded set $B$ since our models are assumed to be globally differentiable in $\R^{d}\times\mathcal{S}$. This compact set only implicitly depends on the training process and the value of $\epsilon$. Based on Lemma~\ref{theorem1} and other common assumptions in stochastic optimization~\citep{bottou2018optimization}, FD-SSM converges to a stationary point of SSM, as stated below:

\begin{theorem}
(Consistency under SGD) Optimizing $\nabla_{\theta}\mathcal{J}_{\textup{FD-SSM}}(\theta)$ with stochastic gradient descent, then the model parameters $\theta$ will converge to the stationary point of $\mathcal{J}_{\textup{SSM}}(\theta)$ under the conditions including: (\romannumeral 1) the assumptions for general stochastic optimization in~\citet{bottou2018optimization} hold; (\romannumeral 2) the differentiability assumptions in Lemma~\ref{theorem1} hold; (\romannumeral 3) $\epsilon$ decays to zero during training.
\end{theorem}
In the proof, we further show that the conditions (\emph{\romannumeral 1}) and (\emph{\romannumeral 2}) largely overlap, and these assumptions are satisfied by the models described in the remark of Sec.~\ref{3.1remark}. As to the condition (\emph{\romannumeral 3}), we observe that in practice it is enough to set $\epsilon$ be a small constant during training, as shown in our experiments.

\vspace{-0.2cm}
\section{Experiments}
\vspace{-0.2cm}
\label{exp}

In this section, we experiment on a diverse set of generative models, following the default settings in previous work~\citep{li2019annealed,song2019generative,song2019sliced}.\footnote{Our code is provided in \url{https://github.com/taufikxu/FD-ScoreMatching}.} It is worth clarifying that we use the same number of training iterations for our FD methods as their gradient-based counterparts, while we report the time per iteration to exclude the compiling time. More implementation and definition details are in Appendix {\color{red} C.2}.

%\junz{any consideration on choosing this set of models? do they cover a diverse set?}\junz{should make clear somewhere that all the reported time is training time per iteration.}

%\vspace{-0.1cm}
\subsection{Energy-based generative models}
%\vspace{-0.1cm}
\textbf{Deep EBMs} utilize the capacity of neural networks to define unnormalized models. The backbone we use is an 18-layer ResNet~\cite{he2016deep} following~\citet{li2019annealed}. We validate our methods on six datasets including MNIST~\citep{Lecun1998}, Fashion-MNIST~\citep{xiao2017fashion}, CelebA~\citep{liu2015faceattributes}, CIFAR-10~\citep{Krizhevsky2012}, SVHN~\citep{netzer2011reading}, and ImageNet~\citep{deng2009imagenet}. For CelebA and ImageNet, we adopt the officially cropped images and respectively resize to $32\times 32$ and $128\times 128$. The quantitative results on MNIST are given in Table~\ref{tab:ebm_mnist}. As shown, our FD formulations result in $2.9\times$ and $1.7\times$ speedup compared to the gradient-based SSM and DSM, respectively, with consistent SM losses. We simply set $\epsilon=0.1$ to be a constant during training, since we find that the performance of our FD reformulations is insensitive to a wide value range of $\epsilon$. In Fig.~\ref{fig:trace_plot} (a) and (b), we provide the loss curve of DSM / FD-DSM and SSM / FD-SSM w.r.t. time. As seen, FD-DSM can achieve the best model (lowest SM loss) faster, but eventually converges to higher loss compared to DSM. In contrast, when applying FD on SSM-based methods, the improvements are much more significant. This indicates that the random projection trick required by the FD formula is its main downside, which may outweigh the gain on efficiency for low-order computations.

As an additional evaluation of the learned model's performance, we consider two tasks using deep EBMs: the first one is \textbf{out-of-distribution detection}, where we follow previous work~\citep{choi2018waic,nalisnick2019detecting} to use typicality as the detection metric (details in Appendix {\color{red} C.3}), and report the AUC scores~\citep{hendrycks2016baseline} and the training time per iteration in Table~\ref{OOD}; the second one is \textbf{image generation}, where we apply annealed Langevin dynamics~\citep{li2019annealed,nijkamp2019anatomy,welling2011bayesian,xie2016theory} for inference and show the generated samples in 
the left of Fig.~\ref{fig:ebm_generated}.

% Table generated by Excel2LaTeX from sheet 'Sheet1'
\begin{table}[t]
  \centering
  \vspace{-0.cm}
  \caption{Results of the DKEF model on three UCI datasets. We report the negative log-likelihood (NLL) and the exact SM loss on the test set, as well as the training time per iteration. Under each algorithm, we train the DKEF model for 500 epochs with the batch size of 200.}
  \vspace{0.1cm}
    \begin{tabular}{c|ccc|ccc|ccc}
    \Xhline{2\arrayrulewidth}
         \!\!\! \multirow{2}{*}{Algorithm} \!\!\!  & \multicolumn{3}{c|}{Parkinsons} & \multicolumn{3}{c|}{RedWine} & \multicolumn{3}{c}{WhiteWine} \\
          & \!\! NLL \!\!   & \!\! SM loss \!\!   & \!\! Time \!\! & \!\! NLL \!\!   & \!\! SM loss \!\!  & \!\! Time \!\! & \!\! NLL \!\!   & \!\! SM loss \!\!   & \!\! Time \!\! \\
    \hline
    \!\!\! SSM \!\!\!   & \!\! 14.52 \!\! & \!\! $-$123.54 \!\! & \!\! 110 ms \!\! & \!\! 13.34 \!\! & \!\! $-$33.28 \!\! & \!\! 113 ms \!\! & \!\! 14.13 \!\! & \!\! $-$38.43 \!\! & \!\! 105 ms \!\! \\
    
    \!\!\! SSMVR \!\!\! & \!\! 13.26 \!\! & \!\! $-$193.97 \!\! & \!\! 111 ms \!\! & \!\! 13.13 \!\!  & \!\! $-$31.19 \!\! & \!\! 106 ms \!\! & \!\! 13.63 \!\! & \!\! $-$39.42 \!\! & \!\! 111 ms \!\! \\
    
    \!\!\! \textbf{FD-SSM} \!\!\! & \!\! 13.69 \!\! & \!\! $-$138.72 \!\! & \!\! \textbf{82.5 ms} \!\! &  \!\! 13.06 \!\! & \!\! $-$30.34 \!\! & \!\! \textbf{82 ms} \!\!  & \!\! 14.10 \!\! & \!\! $-$32.84 \!\! & \!\! \textbf{81.0 ms} \!\! \\
   \Xhline{2\arrayrulewidth}
    \end{tabular}%
  \label{tab:dkef_uci}%
\end{table}%

\textbf{Deep kernel exponential family~(DKEF)}~\citep{wenliang2019learning} is another unnormalized density estimator in the form of $\log \tilde{p}(x) = f(x) + \log p_0(x)$, with $p_0$ be the base measure, $f(x)$ defined as $\sum_{i=1}^{N}\sum_{j=1}^{N_j}k_i(x, z_j)$, where $N$ is the number of kernels, $k(\cdot, \cdot)$ is the Gaussian kernel function, and ${z_j}$ ($j=0,\cdots,N_j$) are $N_j$ inducing points. The features are extracted using a neural network and the parameters of both the network and the kernel can be learned jointly using SM.
Following the setting in \citet{song2019sliced}, we evaluate on three UCI datasets~\citep{asuncion2007uci} and report the results in Table~\ref{tab:dkef_uci}. As done for SSM, we calculate the tractable solution of the kernel method when training DKEF. The shared calculation leads to a relatively lower speed-up ratio of our FD method compared to the deep EBM case. For the choice of $\epsilon$, we found that the performances are insensitive to $\epsilon$: on the Parkinson dataset, the test NLLs and their corresponding $\epsilon$ are: 14.17$(\epsilon = 0.1)$, 13.51$(\epsilon = 0.05)$, 14.03$(\epsilon = 0.02)$, 14.00$(\epsilon = 0.01)$.

\begin{figure}[t]
\vspace{-0.2cm}
\begin{minipage}[t]{.435\linewidth}
\captionof{table}{Results of deep EBMs on MNIST trained for 300K iterations with the batch size of 64. Here $^\star$ indicates non-parallelized implementation of the FD objectives.}
\vspace{-0.35cm}
  \begin{center}
  \begin{small}
  %\begin{sc}
  \begin{tabular}{c|c|c|c}
    \hline
  \!\!\! Algorithm \!\!\!& SM loss &Time & Mem.\\
    \hline
    DSM   &  \!\!\!  $-9.47\times10^4$ \!\!\!  &   282 ms        &  3.0 G\\
    \!\!\!\textbf{FD-DSM$^\star$} \!\!\!&  \!\!\! $-9.24\times10^4$ \!\!\!   &   191 ms       &  3.2 G\\
    \!\!\!\textbf{FD-DSM} \!\!\!  & \!\!\! $-9.27\times10^4$ \!\!\!    &   \textbf{162 ms}          &  2.7 G\\
    \hline
    \!\!\!SSM\!\!\! & \!\!\!  $-2.97\times 10^7$  \!\!\!  &  673 ms    &      5.1 G\\
    \!\!\!SSMVR \!\!\!  &  \!\!\!  $-3.09\times10^7$ \!\!\!  &   670 ms    &   5.0 G\\
    \!\!\!\textbf{FD-SSM$^\star$}\!\!\! & \!\!\!  $-3.36\times10^7$  \!\!\!  &   276 ms    &   3.7 G \\
    \!\!\!\textbf{FD-SSM} \!\!\!&  \!\!\!  $-3.33\times10^7$ \!\!\!  &   \textbf{230 ms} &       3.4 G\\
    \hline
    \end{tabular}%
  \label{tab:ebm_mnist}%
 % \end{sc}
  \end{small}
  \end{center}
  \end{minipage}
  \hspace{0.55cm}
\begin{minipage}[t]{.52\linewidth}
\captionof{table}{Results of the NICE model trained for 100 epochs with the batch size of 128 on MNIST. Here $^{\dagger}$ indicates $\sigma\!=\!0.1$~\citep{song2019sliced} and $^{\dagger\dagger}$ indicates $\sigma\!=\!1.74$~\citep{saremi2018deep}.}
\vspace{-0.2cm}
  \begin{center}
  \begin{small}
  %\begin{sc}
  \begin{tabular}{c|c|c|c}
    \hline
   \!\!\!\!\!Algorithm\!\!\! &\!\!\!\!SM loss\!\!\! &\!\!\! NLL \!\!\! & \!\!\! Time \!\!\!\\
    \hline
   \!\! \!\!\!Approx BP \!\!\!  &  \!\!\! $-$2530 $\pm$ $617$ \!\!\!  & \!\!\!  1853 $\pm$ 819 \!\!\! & \!\!\! 55.3 ms \!\!\!\\
   \!\!\!\!\! CP \!\!\! & \!\!\! $-$2049 $\pm$ 630 \!\!\!  &  \!\!\!  1626 $\pm$ 269 \!\!\!  & \!\!\!  73.6 ms \!\!\!\\
   \!\! \!\!\!\!DSM$^{\dagger}$ \!\!\!  &  \!\!\! $-$2820 $\pm$ 825 \!\!\!  &  \!\!\! 3398 $\pm$ 1343 \!\!\!  &  \!\!\!  35.8 ms \!\!\!\\
  \!\!\!\!\!\! DSM$^{\dagger\dagger}$ \!\!\!& \!\!\! $-$180 $\pm$ 50 \!\!\!  &  \!\!\! 3764 $\pm$ 1583 \!\!\!  &\!\!\! 37.2 ms \!\!\!\\
  \!\!\!\!\!SSM \!\!\!&\!\!\! $-$2182 $\pm$ 269 \!\!\!& \!\!\! 2579 $\pm$ 945 \!\!\!  & \!\!\!  59.6 ms \!\!\! \\
   \!\!\!\!\!SSMVR \!\!\!&\!\!\! $-$4943 $\pm$ 3191 \!\!\! & \!\!\!  6234 $\pm$ 3782 \!\!\!  & \!\!\!  61.7 ms \!\!\!\\
   \!\!\!\!\!\textbf{FD-SSM} \!\!\!&\!\!\! $-$2425 $\pm$ 100 \!\!\!&  \!\!\!  1647 $\pm$ 306 \!\!\!  &  \!\!\!  \textbf{26.4 ms} \!\!\!\\
   \hline
   \!\!\!\!\! MLE \!\!\!  &  \!\!\!  $-$1236 $\pm$ 525 \!\!\!  &  \!\!\!  791 $\pm$ 14 & \!\!\! 24.3 ms \!\!\!\\
    \hline
    \end{tabular}%
  \label{tab:nice_mnist}%
 % \end{sc}
  \end{small}
  \end{center}
\end{minipage}
  \vspace{-0.3cm}
  \end{figure}

\vspace{-0.15cm}
\subsection{Flow-based generative models}
\vspace{-0.15cm}
In addition to the unnormalized density estimators, SM methods can also be applied to flow-based models, whose log-likelihood functions are tractable and can be directly trained with MLE. Following~\citet{song2019sliced}, we adopt the NICE~\citep{dinh2014nice} model and train it by minimizing the Fisher divergence using different approaches including approximate back-propagation (Approx BP)~\citep{kingma2010regularized} and curvature propagation (CP)~\citep{martens2012estimating}. As in Table~\ref{tab:nice_mnist}, FD-SSM achieves consistent results compared to SSM, while the training time is nearly comparable with the direct MLE, due to parallelization.
The results are averaged over 5 runs except the SM based methods which are averaged over 10 runs. Howev the variance is still large. We hypothesis that it is because the numerical stability of the baseline methods are relatively poor.
In contrast, the variance of FD-SSM on the SM loss is much smaller, which shows better numerical stability of the shallower computational graphs induced by the FD decomposition.

%\vspace{-0.15cm}
\subsection{Latent variable models with implicit encoders}
%\vspace{-0.15cm}
SM methods can be also used in score estimation~\citep{li2017gradient,sasaki2014clustering,strathmann2015gradient}. One particular application is on VAE~\citep{kingma2013auto} / WAE~\citep{tolstikhin2017wasserstein} with implicit encoders, where the gradient of the entropy term in the ELBO w.r.t. model parameters can be estimated (more details can be found in \citet{song2019sliced} and \citet{shi2018spectral}). We follow~\citet{song2019sliced} to evaluate VAE / WAE on both the MNIST and CelebA datasets using both SSMVR and FD-SSMVR. We report the results in Table \ref{tab:vae_wae}. The reported training time only consists of the score estimation part, i.e., training the score model. As expected, the FD reformulation can improve computational efficiency without sacrificing the performance. The discussions concerned with other applications on the latent variable models can be found in Appendix~{\color{red} B.6}.

%\vspace{-0.15cm}
\subsection{Score-based generative models}
\begin{figure}[t]
    \centering
    % \vspace{-0.3cm}
    \includegraphics[width=.9\textwidth]{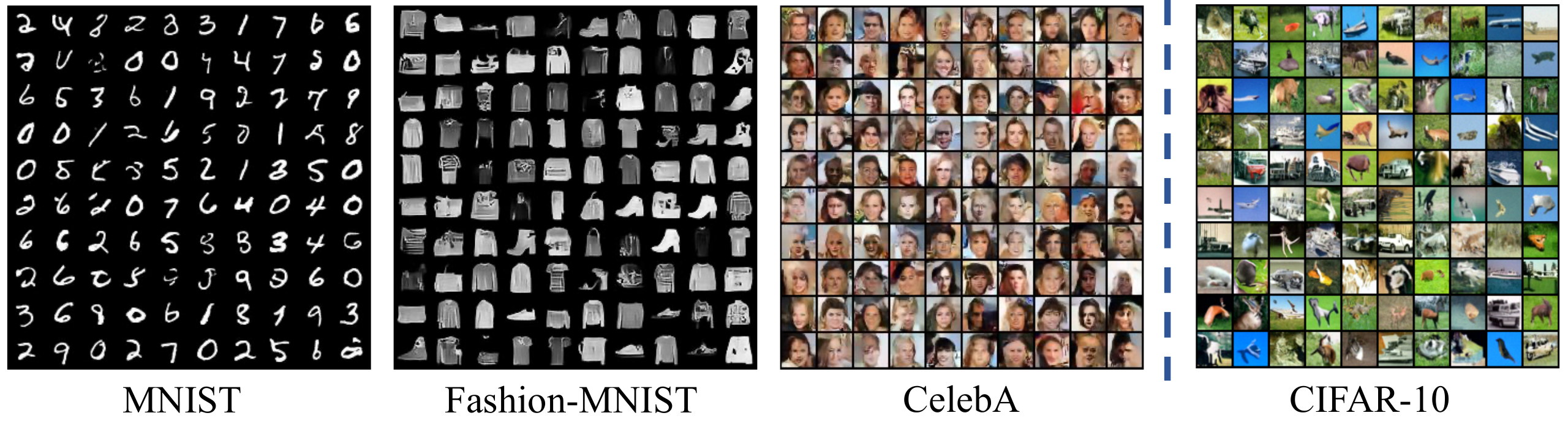}
    \vspace{-0.15cm}
    \caption{\emph{Left.} The generated samples from deep EBMs trained by FD-DSM on MNIST, Fashion-MNIST and CelebA; \emph{Right.} The generated samples from NCSN trained by FD-SSMVR on CIFAR-10.}
    \vspace{-0.25cm}
    \label{fig:ebm_generated}
\end{figure}

%\vspace{-0.15cm}
The noise conditional score network (NCSN)~\citep{song2019generative} trains a single score network $s_{\theta}(x,\sigma)$ to estimate the scores corresponding to all noise levels of $\sigma$. The noise level $\{\sigma_{i}\}_{i\in[10]}$ is a geometric sequence with $\sigma_{1}=1$ and $\sigma_{10}=0.01$. When using the annealed Langevin dynamics for image generation, the number of iterations under each noise level is $100$ with a uniform noise as the initial sample. As to the training approach of NCSN, \citet{song2019generative} mainly use DSM to pursue state-of-the-art performance, while we use SSMVR to demonstrate the efficiency of our FD reformulation. We train the models on the CIFAR-10 dataset with the batch size of $128$ and compute the FID scores~\citep{heusel2017gans} on $50,000$ generated samples. We report the results in Table~\ref{tab:ncsn} and provide the generated samples in the right panel of Fig.~\ref{fig:ebm_generated}. We also provide a curve in Fig.~\ref{fig:trace_plot} (c) showing the FID scores (on 1,000 samples) during training. As seen, our FD methods can effectively learn different generative models.

% The results are given in Table~\ref{tab:vae_wae}. The reported training time only consists of the score estimation part, i.e., training the score model. As expected, the FD reformulation can improve computational efficiency without sacrificing the performance.

\begin{figure}[t]
\begin{minipage}[t]{.66\linewidth}
\captionof{table}{Results of the out-of-distribution detection on deep EBMs. Training time per iteration and AUC scores ($M\!=\!2$ in typicality).}
%The datasets in each column and row indicate training and test sets, respectively.}
\vspace{-0.15cm}
  \centering
  \begin{tabular}{c|c|c|c|c|c}
    \Xhline{2\arrayrulewidth}
    \!\!\!  {Dataset} \!\!\! & \!\!\!\! {Algorithm} \!\!\!\! & Time & \!\!\! {SVHN} \!\!\! & \!\!\! {CIFAR} \!\!\! & \!\!\! {ImageNet} \!\!\! \\
    \hline

    \!\!\! \multirow{2}{*}{SVHN} \!\!\! & \!\!\!\! DSM \!\!\!\! & 673 ms & 0.49 & 1.00  & 0.99 \\
    & \!\!\!\! \textbf{FD-DSM} \!\!\!\! & \textbf{305 ms} & 0.50 & 1.00 & 1.00 \\
    \hline
   \!\!\! \multirow{2}{*}{CIFAR} \!\!\!& \!\!\!\! DSM \!\!\!\! & 635 ms & 0.91 & 0.49 & 0.79 \\
    & \!\!\!\! \textbf{FD-DSM} \!\!\!\! & \textbf{311 ms} & 0.92 & 0.51 & 0.81\\
    \hline
    \!\!\! \multirow{2}{*}{ImageNet} \!\!\!& \!\! DSM \!\! & 1125 ms & 0.95 & 0.87 & 0.49\\
    & \!\!\!\! \textbf{FD-DSM} \!\!\!\! & \textbf{713 ms} & 0.95 & 0.89 & 0.49 \\
    \Xhline{2\arrayrulewidth}
    \end{tabular}%
    \label{OOD}
%     \begin{tabular}{c|c|cc|cc}
%     \toprule
%          \multirow{2}{*}{Model} & \multirow{2}{*}{Algorithm}& \multicolumn{2}{c|}{MNIST} & \multicolumn{2}{c}{CelebA} \\
% %\cline{3-6}
%          &  & NLL &  Time & FID &  Time \\
%     \midrule
%     \multirow{2}{*}{VAE}& SSMVR & 89.58 & 5.04 ms & 62.76 & 14.9 ms \\
%     &\textbf{FD-SSMVR} & 88.96 & \textbf{3.98 ms} & 64.85 & \textbf{9.38 ms} \\
%     \midrule
%     \multirow{2}{*}{WAE} & SSMVR & 90.45 &   0.55 ms    &   54.28    &  1.30 ms\\
%     & \textbf{FD-SSMVR} & 90.66 &   \textbf{0.39 ms}    &     54.67  &  \textbf{0.81 ms}\\
%     \bottomrule
%     \end{tabular}
    % \label{tab:vae_wae}
  \end{minipage}
  \hspace{0.3cm}
\begin{minipage}[t]{.32\linewidth}
\vspace{-0.cm}
\captionof{table}{Results of the NCSN model trained for 200K iterations with 128 batch size on CIFAR-10. We report time per iteration and the FID scores.}
\vspace{0.cm}
  \begin{center}
  \begin{small}
  %\begin{sc}
  \begin{tabular}{c|c|c|c}
   \Xhline{2\arrayrulewidth}
    \!\!\! Algorithm \!\!\! & \!\!\! FID \!\!\! & \!\!\! Time \!\!\! & \!\!\! Mem. \!\!\!\!\! \\
    \hline
    \!\!\!\!\! SSMVR \!\!\!\!\! & \!\!\! 41.2 \!\!\!  & \!\!\! 865 ms \!\!\! & \!\!\! 6.4 G \!\!\!\!\! \\
    \!\!\!\!\! \textbf{FD-SSMVR} \!\!\!\!\! & \!\!\! \textbf{39.5} \!\!\!  & \!\!\! \textbf{575 ms} \!\!\! & \!\!\! 5.5 G \!\!\!\!\! \\
    \Xhline{2\arrayrulewidth}

      \end{tabular}
  \label{tab:ncsn}%
 % \end{sc}
  \end{small}
  \end{center}
\end{minipage}
  \vspace{-0.05cm}
  \end{figure}

\begin{table}[t]
  \centering
%   \vspace{-0.6cm}
  \caption{The results of training implicit encoders for VAE and WAE on the MNIST and CelebA datasets. The models are trained for 100K iterations with the batch size of 128.}
  \vspace{0.2cm}
    \begin{tabular}{c|c|cc|cc}
    \toprule
         \multirow{2}{*}{Model} & \multirow{2}{*}{Algorithm}& \multicolumn{2}{c|}{MNIST} & \multicolumn{2}{c}{CelebA} \\
%\cline{3-6}
         &  & NLL &  Time & FID &  Time \\
    \midrule
    \multirow{2}{*}{VAE}& SSMVR & 89.58 & 5.04 ms & 62.76 & 14.9 ms \\
    &\textbf{FD-SSMVR} & 88.96 & \textbf{3.98 ms} & 64.85 & \textbf{9.38 ms} \\
    \midrule
    \multirow{2}{*}{WAE} & SSMVR & 90.45 &   0.55 ms    &   54.28    &  1.30 ms\\
    & \textbf{FD-SSMVR} & 90.66 &   \textbf{0.39 ms}    &     54.67  &  \textbf{0.81 ms}\\
    \bottomrule
    \end{tabular}
  \label{tab:vae_wae}%
%   \vspace{-0.3cm}
\end{table}%

%\vspace{-0.25cm}
\section{Related work}
%\vspace{-0.25cm}
%\se{possibly mention numerical analysis, where these techniques are common}
In numerical analysis, the FD approaches play a central role in solving differential equations~\citep{stoer2013introduction}. In machine learning, there have been related efforts devoted to leveraging the FD forms, either explicitly or implicitly. For a general scalar function $\mathcal{L}(x)$, we denote $H(x)$ as the Hessian matrix, $J(x)$ as the gradient, and $\sigma$ be a small value. \citet{lecun1993efficient} introduces a row-wise approximation of Hessian matrix as $H_{k}(x)\approx\frac{1}{\sigma}(J(x+\sigma\bm{e}_{k})-J(x))$, where $H_{k}$ represents the $k$-th row of Hessian matrix and $\bm{e}_{k}$ is the $k$-th Euclidean basis vector. \citet{rifai2011higher} provide a FD approximation for the Frobenius norm of Hessian matrix as $\|H(x)\|_{F}^{2}\approx \frac{1}{\sigma^2}\mathbb{E}[\|J(x+v)-J(x)\|_{2}^{2}]$, where $v\sim\mathcal{N}(0,\sigma^{2}I)$ and the formulas is used to regularize the unsupervised auto-encoders. \citet{moller1990scaled} approximates the Hessian-vector product $H(x)v$ by calculating the 
directional FD as $H(x)v\approx\frac{1}{\sigma}(J(x+\sigma v)-J(x))$. Compared to our work, these previous methods mainly use the first-order terms $J(x)$ to approximate the second-order terms of $H(x)$, while we utilize the linear combinations of the original function $\mathcal{L}(x)$ to estimate high-order terms that exist in the Taylor's expansion, e.g., $v^{\top}H(x)v$.

As to the more implicit connections to FD, the minimum probability flow (MPF)~\citep{sohl2011new} is a method for parameter estimation in probabilistic models. It is demonstrated that MPF can be connected to SM by a FD reformulation, where we provide a concise derivation in Appendix {\color{red} B.7}. The noise-contrastive estimation (NCE)~\citep{gutmann2010noise} train the unnormalized models by comparing the model distribution $p_{\theta}(x)$ with a noise distribution $p_{n}(x)$. It is proven that when we choose $p_{n}(x)=p_{\text{data}}(x+v)$ with a small vector $v$, i.e., $\|v\|=\epsilon$, the NCE objective can be equivalent to a FD approximation for the SSM objective as to an $o(1)$ approximation error rate after scaling~\citep{song2019sliced}. In contrast, our FD-SSM method can achieve $o(\epsilon)$ approximation error with the same computational cost as NCE.

%As to the optimization methods, the variants of SGD like the momentum-based~\citep{qian1999momentum,sutskever2013importance} and the adaptive-based~\citep{dinh2017sharp,Kingma2014} ones can be seen as implicitly computing FD approximations for the diagonal entries of the Hessian matrix~\citep{lecun2012efficient}.

\begin{figure}[t]
    \centering
    \vspace{-0.3cm}
    \includegraphics[width=1.\textwidth]{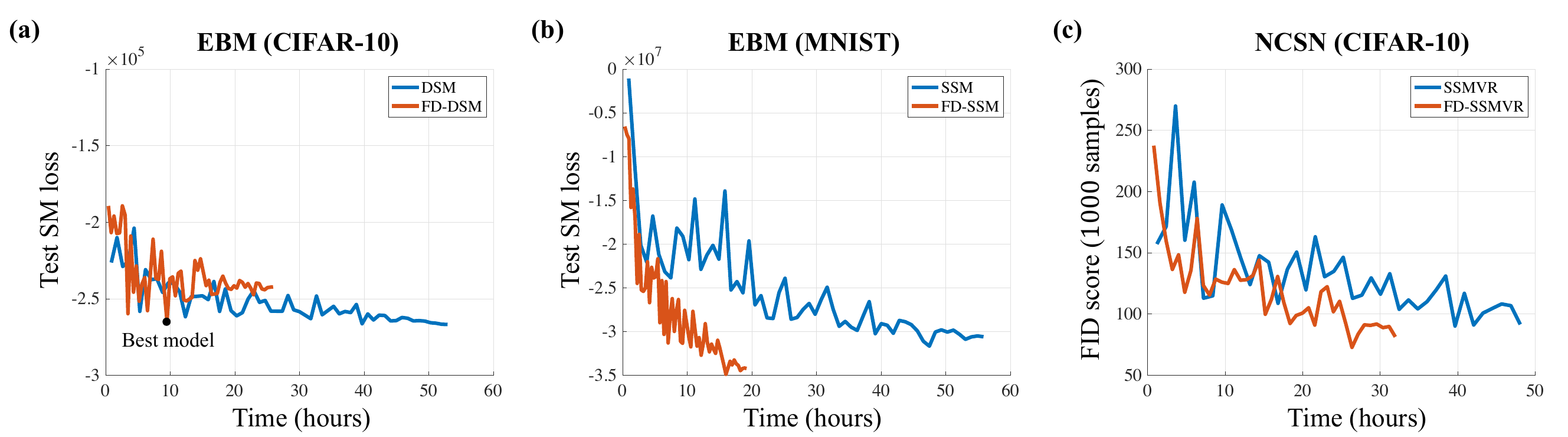}
    %\vspace{-0.15cm}
    \caption{\textbf{(a)} Loss for DSM and FD-DSM; \textbf{(b)} Loss for SSM and FD-SSM; \textbf{(c)} FID scores on 1,000 samples (higher than those reported on 50,000 samples in Table~\ref{tab:ncsn}) for SSMVR and FD-SSMVR.}
    \vspace{-0.1cm}
    \label{fig:trace_plot}
\end{figure}

\vspace{-0.05cm}
\section{Conclusion}
\vspace{-0.05cm}
We propose to reformulate existing gradient-based SM methods using finite difference (FD), and theoretically and empirically demonstrate the consistency and computational efficiency of the FD-based training objectives. In addition to generative modeling, our generic FD decomposition can potentially be used in other applications involving higher-order derivatives. However, the price paid for this significant efficiency is that we need to work on the projected function in a certain direction, e.g., in DSM we need to first convert it into the slice Wasserstein distance and then apply the FD reformulation. This raises a trade-off between efficiency and variance in some cases. 
%Still, this work provides a successful case on using the FD forms in machine learning methods, which is insightful for future exploration.

%\clearpage

\vspace{-0.05cm}
\section*{Broader Impact}
\vspace{-0.05cm}
This work proposes an efficient way to learn generative models and does not have a direct impact on society. However, by reducing the computation required for training unnormalized models, it may facilitate large-scale applications of, e.g., EBMs 
to real-world problems, which could have both positive (e.g., anomaly detection and denoising) and negative (e.g., deepfakes) consequences.
%in the practical systems and consequently lead to other potential impacts in the society.

\section*{Acknowledgements}
This work was supported by the National Key Research and Development Program of China (No.2017YFA0700904), NSFC Projects (Nos. 61620106010, 62076145, U19B2034, U1811461), Beijing Academy of Artificial Intelligence (BAAI), Tsinghua-Huawei Joint Research Program, a grant from Tsinghua Institute for Guo Qiang, Tiangong Institute for Intelligent Computing, and the NVIDIA NVAIL Program with GPU/DGX Acceleration. C. Li was supported by the Chinese postdoctoral innovative talent support program and Shuimu Tsinghua Scholar.

\bibliographystyle{plainnat}
\bibliography{main}

\begin{thebibliography}{82}
\providecommand{\natexlab}[1]{#1}
\providecommand{\url}[1]{\texttt{#1}}
\expandafter\ifx\csname urlstyle\endcsname\relax
  \providecommand{\doi}[1]{doi: #1}\else
  \providecommand{\doi}{doi: \begingroup \urlstyle{rm}\Url}\fi

\bibitem[Arjovsky et~al.(2017)Arjovsky, Chintala, and
  Bottou]{arjovsky2017wasserstein}
Martin Arjovsky, Soumith Chintala, and L{\'e}on Bottou.
\newblock Wasserstein gan.
\newblock \emph{arXiv preprint arXiv:1701.07875}, 2017.

\bibitem[Asuncion and Newman(2007)]{asuncion2007uci}
Arthur Asuncion and David Newman.
\newblock Uci machine learning repository, 2007.

\bibitem[Bao et~al.(2020)Bao, Li, Xu, Su, Zhu, and Zhang]{bao-bi}
Fan Bao, Chongxuan Li, Kun Xu, Hang Su, Jun Zhu, and Bo~Zhang.
\newblock Bi-level score matching for learning energy-based latent variable
  models.
\newblock In \emph{https://arxiv.org/abs/2010.07856}, 2020.

\bibitem[Bottou et~al.(2018)Bottou, Curtis, and
  Nocedal]{bottou2018optimization}
L{\'e}on Bottou, Frank~E Curtis, and Jorge Nocedal.
\newblock Optimization methods for large-scale machine learning.
\newblock \emph{Siam Review}, 60\penalty0 (2):\penalty0 223--311, 2018.

\bibitem[Brunel and Nadal(1998)]{brunel1998mutual}
Nicolas Brunel and Jean-Pierre Nadal.
\newblock Mutual information, fisher information, and population coding.
\newblock \emph{Neural computation}, 10\penalty0 (7):\penalty0 1731--1757,
  1998.

\bibitem[Burda et~al.(2015)Burda, Grosse, and
  Salakhutdinov]{burda2015importance}
Yuri Burda, Roger Grosse, and Ruslan Salakhutdinov.
\newblock Importance weighted autoencoders.
\newblock \emph{arXiv preprint arXiv:1509.00519}, 2015.

\bibitem[Choi et~al.(2018)Choi, Jang, and Alemi]{choi2018waic}
Hyunsun Choi, Eric Jang, and Alexander~A Alemi.
\newblock Waic, but why? generative ensembles for robust anomaly detection.
\newblock \emph{arXiv preprint arXiv:1810.01392}, 2018.

\bibitem[Deng et~al.(2009)Deng, Dong, Socher, Li, Li, and
  Fei-Fei]{deng2009imagenet}
Jia Deng, Wei Dong, Richard Socher, Li-Jia Li, Kai Li, and Li~Fei-Fei.
\newblock Imagenet: A large-scale hierarchical image database.
\newblock In \emph{IEEE Conference on Computer Vision and Pattern Recognition
  (CVPR)}, 2009.

\bibitem[Dinh et~al.(2014)Dinh, Krueger, and Bengio]{dinh2014nice}
Laurent Dinh, David Krueger, and Yoshua Bengio.
\newblock Nice: Non-linear independent components estimation.
\newblock \emph{arXiv preprint arXiv:1410.8516}, 2014.

\bibitem[Du and Mordatch(2019)]{du2019implicit}
Yilun Du and Igor Mordatch.
\newblock Implicit generation and modeling with energy based models.
\newblock In \emph{Advances in Neural Information Processing Systems
  (NeurIPS)}, 2019.

\bibitem[Dumoulin et~al.(2017)Dumoulin, Shlens, and
  Kudlur]{dumoulin2016learned}
Vincent Dumoulin, Jonathon Shlens, and Manjunath Kudlur.
\newblock A learned representation for artistic style.
\newblock In \emph{International Conference on Learning Representations
  (ICLR)}, 2017.

\bibitem[Goodfellow et~al.(2014)Goodfellow, Pouget-Abadie, Mirza, Xu,
  Warde-Farley, Ozair, Courville, and Bengio]{goodfellow2014generative}
Ian Goodfellow, Jean Pouget-Abadie, Mehdi Mirza, Bing Xu, David Warde-Farley,
  Sherjil Ozair, Aaron Courville, and Yoshua Bengio.
\newblock Generative adversarial nets.
\newblock In \emph{Advances in neural information processing systems
  (NeurIPS)}, pages 2672--2680, 2014.

\bibitem[Grathwohl et~al.(2020)Grathwohl, Wang, Jacobsen, Duvenaud, Norouzi,
  and Swersky]{Grathwohl2020Your}
Will Grathwohl, Kuan-Chieh Wang, Joern-Henrik Jacobsen, David Duvenaud,
  Mohammad Norouzi, and Kevin Swersky.
\newblock Your classifier is secretly an energy based model and you should
  treat it like one.
\newblock In \emph{International Conference on Learning Representations
  (ICLR)}, 2020.

\bibitem[Griewank(1993)]{griewank1993some}
Andreas Griewank.
\newblock Some bounds on the complexity of gradients, jacobians, and hessians.
\newblock In \emph{Complexity in numerical optimization}, pages 128--162. World
  Scientific, 1993.

\bibitem[Griewank and Walther(2008)]{griewank2008evaluating}
Andreas Griewank and Andrea Walther.
\newblock \emph{Evaluating derivatives: principles and techniques of
  algorithmic differentiation}, volume 105.
\newblock Siam, 2008.

\bibitem[Gutmann and Hyv{\"a}rinen(2010)]{gutmann2010noise}
Michael Gutmann and Aapo Hyv{\"a}rinen.
\newblock Noise-contrastive estimation: A new estimation principle for
  unnormalized statistical models.
\newblock In \emph{International Conference on Artificial Intelligence and
  Statistics (AISTATS)}, pages 297--304, 2010.

\bibitem[Haarnoja et~al.(2017)Haarnoja, Tang, Abbeel, and
  Levine]{haarnoja2017reinforcement}
Tuomas Haarnoja, Haoran Tang, Pieter Abbeel, and Sergey Levine.
\newblock Reinforcement learning with deep energy-based policies.
\newblock In \emph{International Conference on Machine Learning (ICML)}, 2017.

\bibitem[He et~al.(2019)He, Spokoyny, Neubig, and
  Berg-Kirkpatrick]{he2019lagging}
Junxian He, Daniel Spokoyny, Graham Neubig, and Taylor Berg-Kirkpatrick.
\newblock Lagging inference networks and posterior collapse in variational
  autoencoders.
\newblock \emph{arXiv preprint arXiv:1901.05534}, 2019.

\bibitem[He et~al.(2016)He, Zhang, Ren, and Sun]{he2016deep}
Kaiming He, Xiangyu Zhang, Shaoqing Ren, and Jian Sun.
\newblock Deep residual learning for image recognition.
\newblock In \emph{IEEE Conference on Computer Vision and Pattern Recognition
  (CVPR)}, pages 770--778, 2016.

\bibitem[Hendrycks and Gimpel(2016)]{hendrycks2016baseline}
Dan Hendrycks and Kevin Gimpel.
\newblock A baseline for detecting misclassified and out-of-distribution
  examples in neural networks.
\newblock \emph{arXiv preprint arXiv:1610.02136}, 2016.

\bibitem[Heusel et~al.(2017)Heusel, Ramsauer, Unterthiner, Nessler, and
  Hochreiter]{heusel2017gans}
Martin Heusel, Hubert Ramsauer, Thomas Unterthiner, Bernhard Nessler, and Sepp
  Hochreiter.
\newblock Gans trained by a two time-scale update rule converge to a local nash
  equilibrium.
\newblock In \emph{Advances in Neural Information Processing Systems
  (NeurIPS)}, pages 6626--6637, 2017.

\bibitem[Hinton(2002)]{hinton2002training}
Geoffrey~E Hinton.
\newblock Training products of experts by minimizing contrastive divergence.
\newblock \emph{Neural computation}, 14\penalty0 (8):\penalty0 1771--1800,
  2002.

\bibitem[Hyv{\"a}rinen(2005)]{hyvarinen2005estimation}
Aapo Hyv{\"a}rinen.
\newblock Estimation of non-normalized statistical models by score matching.
\newblock \emph{Journal of Machine Learning Research (JMLR)}, 6\penalty0
  (Apr):\penalty0 695--709, 2005.

\bibitem[Isaacson and Keller(2012)]{isaacson2012analysis}
Eugene Isaacson and Herbert~Bishop Keller.
\newblock \emph{Analysis of numerical methods}.
\newblock Courier Corporation, 2012.

\bibitem[Johnson(2004)]{johnson2004information}
Oliver Johnson.
\newblock \emph{Information theory and the central limit theorem}.
\newblock World Scientific, 2004.

\bibitem[Kingma and Ba(2014)]{kingma2014adam}
Diederik~P Kingma and Jimmy Ba.
\newblock Adam: A method for stochastic optimization.
\newblock \emph{arXiv preprint arXiv:1412.6980}, 2014.

\bibitem[Kingma and Welling(2014)]{kingma2013auto}
Diederik~P Kingma and Max Welling.
\newblock Auto-encoding variational bayes.
\newblock In \emph{International Conference on Learning Representations
  (ICLR)}, 2014.

\bibitem[Kingma and Cun(2010)]{kingma2010regularized}
Durk~P Kingma and Yann~L Cun.
\newblock Regularized estimation of image statistics by score matching.
\newblock In \emph{Advances in neural information processing systems
  (NeurIPS)}, pages 1126--1134, 2010.

\bibitem[Kingma and Dhariwal(2018)]{kingma2018glow}
Durk~P Kingma and Prafulla Dhariwal.
\newblock Glow: Generative flow with invertible 1x1 convolutions.
\newblock In \emph{Advances in Neural Information Processing Systems
  (NeurIPS)}, 2018.

\bibitem[K{\"o}nigsberger(2004)]{konigsberger2004analysis}
Konrad K{\"o}nigsberger.
\newblock Analysis 2 springer verlag, 2004.

\bibitem[Krizhevsky and Hinton(2009)]{Krizhevsky2012}
Alex Krizhevsky and Geoffrey Hinton.
\newblock Learning multiple layers of features from tiny images.
\newblock Technical report, Citeseer, 2009.

\bibitem[Kuleshov and Ermon(2017)]{kuleshov2017neural}
Volodymyr Kuleshov and Stefano Ermon.
\newblock Neural variational inference and learning in undirected graphical
  models.
\newblock In \emph{Advances in Neural Information Processing Systems
  (NeurIPS)}, 2017.

\bibitem[Kumar et~al.(2019)Kumar, Ozair, Goyal, Courville, and
  Bengio]{kumar2019maximum}
Rithesh Kumar, Sherjil Ozair, Anirudh Goyal, Aaron Courville, and Yoshua
  Bengio.
\newblock Maximum entropy generators for energy-based models.
\newblock \emph{arXiv preprint arXiv:1901.08508}, 2019.

\bibitem[Kurach et~al.(2018)Kurach, Lucic, Zhai, Michalski, and
  Gelly]{kurach2018large}
Karol Kurach, Mario Lucic, Xiaohua Zhai, Marcin Michalski, and Sylvain Gelly.
\newblock A large-scale study on regularization and normalization in gans.
\newblock \emph{arXiv preprint arXiv:1807.04720}, 2018.

\bibitem[LeCun(1993)]{lecun1993efficient}
Yann LeCun.
\newblock Efficient learning and second-order methods.
\newblock \emph{A tutorial at NIPS}, 93:\penalty0 61, 1993.

\bibitem[LeCun et~al.(1998)LeCun, Bottou, Bengio, and Haffner]{Lecun1998}
Yann LeCun, L{\'e}on Bottou, Yoshua Bengio, and Patrick Haffner.
\newblock Gradient-based learning applied to document recognition.
\newblock \emph{Proceedings of the IEEE}, 86\penalty0 (11):\penalty0
  2278--2324, 1998.

\bibitem[LeCun et~al.(2006)LeCun, Chopra, Hadsell, Ranzato, and
  Huang]{lecun2006tutorial}
Yann LeCun, Sumit Chopra, Raia Hadsell, M~Ranzato, and F~Huang.
\newblock A tutorial on energy-based learning.
\newblock \emph{Predicting structured data}, 1\penalty0 (0), 2006.

\bibitem[Li and Turner(2018)]{li2017gradient}
Yingzhen Li and Richard~E Turner.
\newblock Gradient estimators for implicit models.
\newblock In \emph{International Conference on Learning Representations
  (ICLR)}, 2018.

\bibitem[Li et~al.(2019)Li, Chen, and Sommer]{li2019annealed}
Zengyi Li, Yubei Chen, and Friedrich~T Sommer.
\newblock Annealed denoising score matching: Learning energy-based models in
  high-dimensional spaces.
\newblock \emph{arXiv preprint arXiv:1910.07762}, 2019.

\bibitem[Lian et~al.(2015)Lian, Huang, Li, and Liu]{lian2015asynchronous}
Xiangru Lian, Yijun Huang, Yuncheng Li, and Ji~Liu.
\newblock Asynchronous parallel stochastic gradient for nonconvex optimization.
\newblock In \emph{Advances in Neural Information Processing Systems
  (NeurIPS)}, pages 2737--2745, 2015.

\bibitem[Lin et~al.(2017)Lin, Milan, Shen, and Reid]{lin2017refinenet}
Guosheng Lin, Anton Milan, Chunhua Shen, and Ian Reid.
\newblock Refinenet: Multi-path refinement networks for high-resolution
  semantic segmentation.
\newblock In \emph{IEEE International Conference on Computer Vision (CVPR)},
  pages 1925--1934, 2017.

\bibitem[Liu et~al.(2016)Liu, Lee, and Jordan]{liu2016kernelized}
Qiang Liu, Jason Lee, and Michael Jordan.
\newblock A kernelized stein discrepancy for goodness-of-fit tests.
\newblock In \emph{International Conference on Machine Learning (ICML)}, 2016.

\bibitem[Liu et~al.(2015)Liu, Luo, Wang, and Tang]{liu2015faceattributes}
Ziwei Liu, Ping Luo, Xiaogang Wang, and Xiaoou Tang.
\newblock Deep learning face attributes in the wild.
\newblock In \emph{Proceedings of International Conference on Computer Vision
  (ICCV)}, December 2015.

\bibitem[Luo et~al.(2020)Luo, Beatson, Norouzi, Zhu, Duvenaud, Adams, and
  Chen]{Luo2020SUMO}
Yucen Luo, Alex Beatson, Mohammad Norouzi, Jun Zhu, David Duvenaud, Ryan~P.
  Adams, and Ricky T.~Q. Chen.
\newblock Sumo: Unbiased estimation of log marginal probability for latent
  variable models.
\newblock In \emph{International Conference on Learning Representations
  (ICLR)}, 2020.

\bibitem[Martens et~al.(2012)Martens, Sutskever, and
  Swersky]{martens2012estimating}
James Martens, Ilya Sutskever, and Kevin Swersky.
\newblock Estimating the hessian by back-propagating curvature.
\newblock \emph{arXiv preprint arXiv:1206.6464}, 2012.

\bibitem[Mescheder et~al.(2018)Mescheder, Geiger, and
  Nowozin]{mescheder2018training}
Lars Mescheder, Andreas Geiger, and Sebastian Nowozin.
\newblock Which training methods for gans do actually converge?
\newblock In \emph{International Conference on Machine Learning (ICML)}, 2018.

\bibitem[Mnih and Hinton(2005)]{mnih2005learning}
Andriy Mnih and Geoffrey Hinton.
\newblock Learning nonlinear constraints with contrastive backpropagation.
\newblock In \emph{International Joint Conference on Neural Networks (IJCNN)},
  volume~2, pages 1302--1307. IEEE, 2005.

\bibitem[M{\o}ller(1990)]{moller1990scaled}
Martin~F M{\o}ller.
\newblock \emph{A scaled conjugate gradient algorithm for fast supervised
  learning}.
\newblock Aarhus University, Computer Science Department, 1990.

\bibitem[Nalisnick et~al.(2019)Nalisnick, Matsukawa, Teh, and
  Lakshminarayanan]{nalisnick2019detecting}
Eric Nalisnick, Akihiro Matsukawa, Yee~Whye Teh, and Balaji Lakshminarayanan.
\newblock Detecting out-of-distribution inputs to deep generative models using
  a test for typicality.
\newblock \emph{arXiv preprint arXiv:1906.02994}, 2019.

\bibitem[Netzer et~al.(2011)Netzer, Wang, Coates, Bissacco, Wu, and
  Ng]{netzer2011reading}
Yuval Netzer, Tao Wang, Adam Coates, Alessandro Bissacco, Bo~Wu, and Andrew~Y
  Ng.
\newblock Reading digits in natural images with unsupervised feature learning.
\newblock In \emph{NIPS Workshop on Deep Learning and Unsupervised Feature
  Learning}, 2011.

\bibitem[Nijkamp et~al.(2019)Nijkamp, Hill, Han, Zhu, and
  Wu]{nijkamp2019anatomy}
Erik Nijkamp, Mitch Hill, Tian Han, Song-Chun Zhu, and Ying~Nian Wu.
\newblock On the anatomy of mcmc-based maximum likelihood learning of
  energy-based models.
\newblock \emph{arXiv preprint arXiv:1903.12370}, 2019.

\bibitem[Oord et~al.(2016)Oord, Kalchbrenner, and Kavukcuoglu]{oord2016pixel}
Aaron van~den Oord, Nal Kalchbrenner, and Koray Kavukcuoglu.
\newblock Pixel recurrent neural networks.
\newblock In \emph{International Conference on Machine Learning (ICML)}, 2016.

\bibitem[Paszke et~al.(2019)Paszke, Gross, Massa, Lerer, Bradbury, Chanan,
  Killeen, Lin, Gimelshein, Antiga, et~al.]{paszke2019pytorch}
Adam Paszke, Sam Gross, Francisco Massa, Adam Lerer, James Bradbury, Gregory
  Chanan, Trevor Killeen, Zeming Lin, Natalia Gimelshein, Luca Antiga, et~al.
\newblock Pytorch: An imperative style, high-performance deep learning library.
\newblock In \emph{Advances in Neural Information Processing Systems
  (NeurIPS)}, pages 8024--8035, 2019.

\bibitem[Poon and Domingos(2011)]{poon2011sum}
Hoifung Poon and Pedro Domingos.
\newblock Sum-product networks: A new deep architecture.
\newblock In \emph{2011 IEEE International Conference on Computer Vision
  Workshops (ICCV Workshops)}, pages 689--690. IEEE, 2011.

\bibitem[Rabin et~al.(2011)Rabin, Peyr{\'e}, Delon, and
  Bernot]{rabin2011wasserstein}
Julien Rabin, Gabriel Peyr{\'e}, Julie Delon, and Marc Bernot.
\newblock Wasserstein barycenter and its application to texture mixing.
\newblock In \emph{International Conference on Scale Space and Variational
  Methods in Computer Vision}, pages 435--446. Springer, 2011.

\bibitem[Rifai et~al.(2011)Rifai, Mesnil, Vincent, Muller, Bengio, Dauphin, and
  Glorot]{rifai2011higher}
Salah Rifai, Gr{\'e}goire Mesnil, Pascal Vincent, Xavier Muller, Yoshua Bengio,
  Yann Dauphin, and Xavier Glorot.
\newblock Higher order contractive auto-encoder.
\newblock In \emph{Joint European Conference on Machine Learning and Knowledge
  Discovery in Databases}, pages 645--660. Springer, 2011.

\bibitem[Robbins and Monro(1951)]{robbins1951stochastic}
Herbert Robbins and Sutton Monro.
\newblock A stochastic approximation method.
\newblock \emph{The annals of mathematical statistics}, pages 400--407, 1951.

\bibitem[Saremi and Hyvarinen(2019)]{saremi2019neural}
Saeed Saremi and Aapo Hyvarinen.
\newblock Neural empirical bayes.
\newblock \emph{Journal of Machine Learning Research (JMLR)}, 20:\penalty0
  1--23, 2019.

\bibitem[Saremi et~al.(2018)Saremi, Mehrjou, Sch{\"o}lkopf, and
  Hyv{\"a}rinen]{saremi2018deep}
Saeed Saremi, Arash Mehrjou, Bernhard Sch{\"o}lkopf, and Aapo Hyv{\"a}rinen.
\newblock Deep energy estimator networks.
\newblock \emph{arXiv preprint arXiv:1805.08306}, 2018.

\bibitem[Sasaki et~al.(2014)Sasaki, Hyv{\"a}rinen, and
  Sugiyama]{sasaki2014clustering}
Hiroaki Sasaki, Aapo Hyv{\"a}rinen, and Masashi Sugiyama.
\newblock Clustering via mode seeking by direct estimation of the gradient of a
  log-density.
\newblock In \emph{Joint European Conference on Machine Learning and Knowledge
  Discovery in Databases}, pages 19--34. Springer, 2014.

\bibitem[Shi et~al.(2018)Shi, Sun, and Zhu]{shi2018spectral}
Jiaxin Shi, Shengyang Sun, and Jun Zhu.
\newblock A spectral approach to gradient estimation for implicit
  distributions.
\newblock In \emph{International Conference on Machine Learning (ICML)}, 2018.

\bibitem[Sohl-Dickstein et~al.(2011)Sohl-Dickstein, Battaglino, and
  DeWeese]{sohl2011new}
Jascha Sohl-Dickstein, Peter~B Battaglino, and Michael~R DeWeese.
\newblock New method for parameter estimation in probabilistic models: minimum
  probability flow.
\newblock \emph{Physical review letters}, 107\penalty0 (22):\penalty0 220601,
  2011.

\bibitem[Song and Ermon(2019)]{song2019generative}
Yang Song and Stefano Ermon.
\newblock Generative modeling by estimating gradients of the data distribution.
\newblock In \emph{Advances in Neural Information Processing Systems
  (NeurIPS)}, pages 11895--11907, 2019.

\bibitem[Song et~al.(2019)Song, Garg, Shi, and Ermon]{song2019sliced}
Yang Song, Sahaj Garg, Jiaxin Shi, and Stefano Ermon.
\newblock Sliced score matching: A scalable approach to density and score
  estimation.
\newblock In \emph{Conference on Uncertainty in Artificial Intelligence (UAI)},
  2019.

\bibitem[Sriperumbudur et~al.(2017)Sriperumbudur, Fukumizu, Gretton,
  Hyv{\"a}rinen, and Kumar]{sriperumbudur2017density}
Bharath Sriperumbudur, Kenji Fukumizu, Arthur Gretton, Aapo Hyv{\"a}rinen, and
  Revant Kumar.
\newblock Density estimation in infinite dimensional exponential families.
\newblock \emph{The Journal of Machine Learning Research (JMLR)}, 18\penalty0
  (1):\penalty0 1830--1888, 2017.

\bibitem[Stoer and Bulirsch(2013)]{stoer2013introduction}
Josef Stoer and Roland Bulirsch.
\newblock \emph{Introduction to numerical analysis}, volume~12.
\newblock Springer Science \& Business Media, 2013.

\bibitem[Strathmann et~al.(2015)Strathmann, Sejdinovic, Livingstone, Szabo, and
  Gretton]{strathmann2015gradient}
Heiko Strathmann, Dino Sejdinovic, Samuel Livingstone, Zoltan Szabo, and Arthur
  Gretton.
\newblock Gradient-free hamiltonian monte carlo with efficient kernel
  exponential families.
\newblock In \emph{Advances in Neural Information Processing Systems
  (NeurIPS)}, pages 955--963, 2015.

\bibitem[Sutherland et~al.(2018)Sutherland, Strathmann, Arbel, and
  Gretton]{sutherland2018efficient}
Dougal Sutherland, Heiko Strathmann, Michael Arbel, and Arthur Gretton.
\newblock Efficient and principled score estimation with nystr{\"o}m kernel
  exponential families.
\newblock In \emph{International Conference on Artificial Intelligence and
  Statistics}, pages 652--660, 2018.

\bibitem[Teh et~al.(2003)Teh, Welling, Osindero, and Hinton]{teh2003energy}
Yee~Whye Teh, Max Welling, Simon Osindero, and Geoffrey~E Hinton.
\newblock Energy-based models for sparse overcomplete representations.
\newblock \emph{Journal of Machine Learning Research (JMLR)}, 4\penalty0
  (Dec):\penalty0 1235--1260, 2003.

\bibitem[Tolstikhin et~al.(2017)Tolstikhin, Bousquet, Gelly, and
  Schoelkopf]{tolstikhin2017wasserstein}
Ilya Tolstikhin, Olivier Bousquet, Sylvain Gelly, and Bernhard Schoelkopf.
\newblock Wasserstein auto-encoders.
\newblock \emph{arXiv preprint arXiv:1711.01558}, 2017.

\bibitem[Vincent(2011)]{vincent2011connection}
Pascal Vincent.
\newblock A connection between score matching and denoising autoencoders.
\newblock \emph{Neural computation}, 23\penalty0 (7):\penalty0 1661--1674,
  2011.

\bibitem[Wang et~al.(2020)Wang, Cheng, Li, Zhu, and Zhang]{wang2020wasserstein}
Ziyu Wang, Shuyu Cheng, Yueru Li, Jun Zhu, and Bo~Zhang.
\newblock A wasserstein minimum velocity approach to learning unnormalized
  models.
\newblock In \emph{International Conference on Artificial Intelligence and
  Statistics (AISTATS)}, 2020.

\bibitem[Welling and Teh(2011)]{welling2011bayesian}
Max Welling and Yee~W Teh.
\newblock Bayesian learning via stochastic gradient langevin dynamics.
\newblock In \emph{International Conference on Machine Learning (ICML)}, 2011.

\bibitem[Wenliang et~al.(2019)Wenliang, Sutherland, Strathmann, and
  Gretton]{wenliang2019learning}
Li~Wenliang, Dougal Sutherland, Heiko Strathmann, and Arthur Gretton.
\newblock Learning deep kernels for exponential family densities.
\newblock In \emph{International Conference on Machine Learning (ICML)}, 2019.

\bibitem[Xiao et~al.(2017)Xiao, Rasul, and Vollgraf]{xiao2017fashion}
Han Xiao, Kashif Rasul, and Roland Vollgraf.
\newblock Fashion-mnist: a novel image dataset for benchmarking machine
  learning algorithms.
\newblock \emph{arXiv preprint arXiv:1708.07747}, 2017.

\bibitem[Xie et~al.(2016)Xie, Lu, Zhu, and Wu]{xie2016theory}
Jianwen Xie, Yang Lu, Song-Chun Zhu, and Yingnian Wu.
\newblock A theory of generative convnet.
\newblock In \emph{International Conference on Machine Learning (ICML)}, 2016.

\bibitem[Xu et~al.(2019)Xu, Li, Wei, Zhu, and Zhang]{xu2019understanding}
Kun Xu, Chongxuan Li, Huanshu Wei, Jun Zhu, and Bo~Zhang.
\newblock Understanding and stabilizing gans' training dynamics with control
  theory.
\newblock \emph{arXiv preprint arXiv:1909.13188}, 2019.

\bibitem[Yu et~al.(2017)Yu, Koltun, and Funkhouser]{yu2017dilated}
Fisher Yu, Vladlen Koltun, and Thomas Funkhouser.
\newblock Dilated residual networks.
\newblock In \emph{IEEE Conference on Computer Vision and Pattern Recognition
  (CVPR)}, pages 472--480, 2017.

\bibitem[Yuan(2014)]{yuan2014rotation}
Feiniu Yuan.
\newblock Rotation and scale invariant local binary pattern based on high order
  directional derivatives for texture classification.
\newblock \emph{Digital Signal Processing}, 26:\penalty0 142--152, 2014.

\bibitem[Zheng et~al.(2015)Zheng, Yang, Liu, Liang, and Li]{zheng2015improving}
Hao Zheng, Zhanlei Yang, Wenju Liu, Jizhong Liang, and Yanpeng Li.
\newblock Improving deep neural networks using softplus units.
\newblock In \emph{International Joint Conference on Neural Networks (IJCNN)},
  pages 1--4. IEEE, 2015.

\bibitem[Zhou et~al.(2020)Zhou, Shi, and Zhu]{zhou2020nonparametric}
Yuhao Zhou, Jiaxin Shi, and Jun Zhu.
\newblock Nonparametric score estimators.
\newblock In \emph{International Conference on Machine Learning (ICML)}, 2020.

\bibitem[Zhu et~al.(2017)Zhu, Park, Isola, and Efros]{zhu2017unpaired}
Jun-Yan Zhu, Taesung Park, Phillip Isola, and Alexei~A Efros.
\newblock Unpaired image-to-image translation using cycle-consistent
  adversarial networks.
\newblock In \emph{IEEE International Conference on Computer Vision (CVPR)},
  pages 2223--2232, 2017.

\end{thebibliography}

\clearpage
\appendix
\section{Proofs}
In this section we provide proofs for the conclusions in the main text.

\subsection{Proof of Lemma 1}
If $\mathcal{L}_{\theta}(x)$ is $T$-times-differentiable at $x$, then according to the general form of multivariate Taylor's theorem~\citep{konigsberger2004analysis}, there is
\begin{equation}
    \mathcal{L}_{\theta}(x+\gamma v)=\sum_{t=0}^{T}\gamma^{t}G_{\theta}^{t}(x,v,\epsilon)+o(\epsilon^{T})\text{, where }G_{\theta}^{t}(x,v,\epsilon)=\left(\frac{\epsilon^t}{t!}\frac{\partial^{t}}{\partial v^{t}}\mathcal{L}_{\theta}(x)\right)\text{.}
\end{equation}
In order to extract the $T$-th component $G_{\theta}^{t}(x,v,\epsilon)$, we arbitrarily select a set of $T+1$ different real values as $\{\gamma_{i}\}_{i\in[T+1]}$, and denote the induced Vandermonde matrix $V$ as
\begin{equation}
    V=\begin{bmatrix}
    1& \gamma_{1}& \gamma_{1}^{2}&\cdots &\gamma_{1}^{T}\\
    \vdots&\vdots&\vdots&\ddots&\vdots\\
    1& \gamma_{(T+1)}& \gamma_{(T+1)}^{2}&\cdots &\gamma_{(T+1)}^{T}
  \end{bmatrix}_{(T+1)\times (T+1)}\text{, and }\bm{\beta}=\begin{pmatrix}
    \beta_{1}\\
    \vdots\\
    \beta_{(T+1)}
  \end{pmatrix}_{(T+1)\times 1}\text{,}
\end{equation}
where $\bm{\beta}$ is the vector of coefficients. The determinant of the Vandermonde matrix $V$ is $\det(V)=\prod_{i<j}(\gamma_{j}-\gamma_{i})\neq 0$ since the values $\gamma_{i}$ are distinct. We consider the linear combination
\begin{equation}
    \sum_{i=1}^{T+1}\beta_{i}\mathcal{L}_{\theta}(x+\gamma_{i} v)=\sum_{t=0}^{T}\phi_{t}G_{\theta}^{t}(x,v,\epsilon)+o(\epsilon^{T})\text{, where }V^{\top}\bm{\beta}=\phi\in\R^{T+1}\text{.}
\end{equation}
To eliminate the term $G_{\theta}^{t}(x,v,\epsilon)$ for any $t<T$ and keep the $T$-th order term $G_{\theta}^{T}(x,v,\epsilon)$, we just need to set the coefficient vector $\bm{\beta}$ be solution of $V^{\top}\bm{\beta}=\bm{e_{(T+1)}}$, where $\bm{e_{(T+1)}}$ is the one-hot vector of the $(T\!+\!1)$-th element. Then we have
\begin{equation}
     \sum_{i=1}^{T+1}\beta_{i}\mathcal{L}_{\theta}(x+\gamma_{i} v)=G_{\theta}^{T}(x,v,\epsilon)+o(\epsilon^{T})\Rightarrow\epsilon^{T}\frac{\partial^{T}}{\partial v^{T}}\mathcal{L}_{\theta}(x)=T!\sum_{i=1}^{T+1}\beta_{i}\mathcal{L}_{\theta}(x+\gamma_{i}v)+o(\epsilon^{T})\text{.}
\end{equation}

\qed

\subsection{Proof of Theorem 1}
Let $K\in \mathbb{N}^{+}$, and $\{\alpha_{k}\}_{k\in[K]}$ be any set of $K$ different positive numbers, $\bm{\beta}=(\beta_{1},\cdots,\beta_{K})\in\R^{K}$ be a coefficient vector. Assuming that $\mathcal{L}_{\theta}(x)$ is $(T\!+\!1)$-times-differentiable at $x$. When $T=2K$ is an even number, we select the coefficient set to be $\{\pm\alpha_{1},\cdots,\pm\alpha_{K}\}$. Then we can construct the linear combination
\begin{equation}
\begin{split}
    &\lambda\mathcal{L}_{\theta}(x)+\frac{1}{2}\sum_{k=1}^{K}\beta_{k}\alpha_{k}^{-2}\left[\mathcal{L}_{\theta}(x+\alpha_{k}v)+\mathcal{L}_{\theta}(x-\alpha_{k}v)\right]\\
    =&\lambda\mathcal{L}_{\theta}(x)+\frac{1}{2}\sum_{k=1}^{K}\beta_{k}\alpha_{k}^{-2}\sum_{t=0}^{T+1}\left(1+(-1)^{t}\right)\alpha_{k}^{t}G_{\theta}^{t}(x,v,\epsilon)+o(\epsilon^{T+1})\\
    =&\lambda\mathcal{L}_{\theta}(x)+\sum_{k=1}^{K}\beta_{k}\alpha_{k}^{-2}\sum_{t=0}^{K}\alpha_{k}^{2t}G_{\theta}^{2t}(x,v,\epsilon)+o(\epsilon^{T+1})\text{,}\\
    \label{even}
\end{split}
\end{equation}
where the second equation holds because $\left(1+(-1)^{t}\right)=0$ for any odd value of $t$. Note that there is $G_{\theta}^{0}(x,v,\epsilon)=\mathcal{L}_{\theta}(x)$, thus in order to eliminate the zero-order term, we let
\begin{equation}
    \lambda=-\sum_{k=1}^{K}\beta_{k}\alpha_{k}^{-2}\text{,}
\end{equation}
and then we can rewrite Eq.~(\ref{even}) as
\begin{equation}
    \begin{split}
    &\frac{1}{2}\sum_{k=1}^{K}\beta_{k}\alpha_{k}^{-2}\left[\mathcal{L}_{\theta}(x+\alpha_{k}v)+\mathcal{L}_{\theta}(x-\alpha_{k}v)-2\mathcal{L}_{\theta}(x)\right]\\
    =&\sum_{k=1}^{K}\beta_{k}\alpha_{k}^{-2}\sum_{t=1}^{K}\alpha_{k}^{2t}G_{\theta}^{2t}(x,v,\epsilon)+o(\epsilon^{T+1})\text{,}\\
    =&\sum_{k=1}^{K}\beta_{k}\sum_{t=0}^{K-1}\alpha_{k}^{2t}G_{\theta}^{2t+2}(x,v,\epsilon)+o(\epsilon^{T+1})\text{,}\\
    \label{even_2}
\end{split}
\end{equation}
Now in Eq.~(\ref{even_2}) we only need to eliminate the term $G_{\theta}^{2t+2}(x,v,\epsilon)$ for $t<K-1$ and keep the term $G_{\theta}^{2K}(x,v,\epsilon)$, i.e., the $T$-th order term.  we define the Vandermonde matrix $V$ generated by $\{\alpha_{1}^{2},\cdots,\alpha_{K}^{2}\}$ as
\begin{equation}
    V=\begin{bmatrix}
    1& \alpha_{1}^{2}& (\alpha_{1}^{2})^{2}&\cdots &(\alpha_{1}^{2})^{K-1}\\
    \vdots&\vdots&\vdots&\ddots&\vdots\\
    1& \alpha_{K}^{2}& (\alpha_{K}^{2})^{2}&\cdots &(\alpha_{K}^{2})^{K-1}
  \end{bmatrix}_{K\times K}\text{.}
\end{equation}
It is easy to know that $V$ is non-singular as long as $\alpha_{k}$ are positive and different. Then if $\bm{\beta}$ is the solution of $V^{\top}\bm{\beta}=\bm{e}_{K}$ is the one-hot vector of the $K$-th element. Then we have
\begin{equation}
\frac{1}{2}\sum_{k=1}^{K}\beta_{k}\alpha_{k}^{-2}\left[\mathcal{L}_{\theta}(x+\alpha_{k}v)+\mathcal{L}_{\theta}(x-\alpha_{k}v)-2\mathcal{L}_{\theta}(x)\right]=G_{\theta}^{T}(x,v,\epsilon)+o(\epsilon^{T+1})\text{.}
\end{equation}

Similarly when $T=2K-1$ is an odd number, we can construct the linear combination
\begin{equation}
\begin{split}
    &\frac{1}{2}\sum_{k=1}^{K}\beta_{k}\alpha_{k}^{-1}\left[\mathcal{L}_{\theta}(x+\alpha_{k}v)-\mathcal{L}_{\theta}(x-\alpha_{k}v)\right]\\
    =&\frac{1}{2}\sum_{k=1}^{K}\beta_{k}\alpha_{k}^{-1}\sum_{t=0}^{T+1}\left(1-(-1)^{t}\right)\alpha_{k}^{t}G_{\theta}^{t}(x,v,\epsilon)+o(\epsilon^{T+1})\\
    =&\sum_{k=1}^{K}\beta_{k}\sum_{t=0}^{K-1}\alpha_{k}^{2t}G_{\theta}^{2t+1}(x,v,\epsilon)+o(\epsilon^{T+1})\text{,}\\
    \label{odd}
\end{split}
\end{equation}
where the second equation holds because $\left(1-(-1)^{t}\right)=0$ for any even value of $t$. Now we only need to eliminate the term $G_{\theta}^{2t+1}(x,v,\epsilon)$ for $t<K-1$ and keep the term $G_{\theta}^{2K-1}(x,v,\epsilon)$, i.e., the $T$-th order term. Then if we still let $\bm{\beta}$ be the solution of $V_{\text{even}}^{\top}\bm{\beta}=\bm{e}_{K}$, we will have
\begin{equation}
\frac{1}{2}\sum_{k=1}^{K}\beta_{k}\alpha_{k}^{-1}\left[\mathcal{L}_{\theta}(x+\alpha_{k}v)-\mathcal{L}_{\theta}(x-\alpha_{k}v)\right]=G_{\theta}^{T}(x,v,\epsilon)+o(\epsilon^{T+1})\text{.}
\end{equation}
\qed

\subsection{Proof of Lemma 2}
We first investigate the gradient $\nabla_{\theta}\mathcal{J}_{\text{FD-SSM}}(x,v;\theta)$, whose elements consist of $\frac{\partial}{\partial \omega}\mathcal{J}_{\text{FD-SSM}}(x,v;\theta)$ for $\omega\in\theta$. Let $\overline{B}_{\epsilon_{0}}$ be the closure of ${B}_{\epsilon_{0}}$, then $\forall(x,\theta)\in B, t\in[-1,1]$, there is $(x+t\cdot v,\theta)\in\overline{B}_{\epsilon_{0}}$ holds for any $v\in\R^{d}, \|v\|_{2}=\epsilon<\epsilon_{0}$. It is easy to verify that $\overline{B}_{\epsilon_{0}}$ is a compact set. Since $\log p_{\theta}(x)$ is four times continuously differentiable in $\overline{B}_{\epsilon_{0}}$, we can obtain
\begin{equation}
\begin{split}
    &\frac{\partial}{\partial\omega}\log p_{\theta}(x+v)\\
    =& \frac{\partial}{\partial\omega}\log p_{\theta}(x)+v^{\top}\nabla_{x}\frac{\partial}{\partial\omega}\log p_{\theta}(x)+\frac{1}{2}v^{\top}\nabla_{x}^{2}\frac{\partial}{\partial\omega}\log p_{\theta}(x)v+\sum_{|\bm{\alpha}|=3}R_{\bm{\alpha}}^{\omega}(x+v)\cdot v^{\bm{\alpha}}\\
    =&\frac{\partial}{\partial\omega}\left[\log p_{\theta}(x)+v^{\top}\nabla_{x}\log p_{\theta}(x)+\frac{1}{2}v^{\top}\nabla_{x}^{2}\log p_{\theta}(x)v\right]+\sum_{|\bm{\alpha}|=3}R_{\bm{\alpha}}^{\omega}(x+v)\cdot v^{\bm{\alpha}}\text{,}
    \label{appendixequ1}
\end{split}
\end{equation}
where $|\bm{\alpha}|=\alpha_{1}+\cdots+\alpha_{d}$, $\bm{\alpha}!=\alpha_{1}!\cdots\alpha_{d}!$, $v^{\bm{\alpha}}=v_{1}^{\alpha_{1}}\cdots v_{d}^{\alpha_{d}}$, and
\begin{equation}
\begin{split}
&R_{\bm{\alpha}}^{\omega}(x+v)=\frac{|\bm{\alpha}|}{\bm{\alpha}!}\int_{0}^{1}(1-t)^{|\bm{\alpha}|-1}D^{\bm{\alpha}}\frac{\partial}{\partial\omega}\log p_{\theta}(x+t\cdot v)dt\text{,}\\
    &D^{\bm{\alpha}}=\frac{\partial^{|\bm{\alpha}|}}{\partial x_{1}^{\alpha_{1}}\cdots \partial x_{d}^{\alpha_{d}}}\text{.}
\end{split}
\label{Ddefinition}
\end{equation}
Due to the continuity of the fourth-order derivatives of $\log p_{\theta}(x)$ on the compact set $\overline{B}_{\epsilon_{0}}$, we can obtain the uniform upper bound for $\forall(x,\theta)\in B, v\in\R^{d}, \|v\|_{2}=\epsilon<\epsilon_{0}$ that
\begin{equation}
    |R_{\bm{\alpha}}^{\omega}(x+v)|\leq U_{\omega}^{+}(\bm{\alpha})\leq\max_{\bm{\alpha}}U_{\omega}^{+}(\bm{\alpha})=U_{\omega}^{+}\text{.}
\end{equation}
So the remainder term in Eq.~(\ref{appendixequ1}) has a upper bound as
\begin{equation}
    \left|\sum_{|\bm{\alpha}|=3}R_{\bm{\alpha}}^{\omega}(x+v)\cdot v^{\bm{\alpha}}\right|<d^{3}\epsilon^{3}U_{\omega}^{+}\text{,}
\end{equation}
where similar results also hold for $x-v$ and we represent the corresponding upper bound as $U_{\omega}^{-}$. Then we further have
\begin{equation}
\begin{split}
    &\frac{\partial}{\partial\omega}\left[\log p_{\theta}(x+v)+\log p_{\theta}(x-v)-2\log p_{\theta}(x)\right]\\
    =&\frac{\partial}{\partial\omega}v^{\top}\nabla_{x}^{2}\log p_{\theta}(x)v+\sum_{|\bm{\alpha}|=3}\left(R_{\bm{\alpha}}^{\omega}(x+v)+R_{\bm{\alpha}}^{\omega}(x-v)\right)\cdot v^{\bm{\alpha}}\text{;}\\
    &\frac{\partial}{\partial\omega}\left[\log p_{\theta}(x+v)-\log p_{\theta}(x-v)\right]\\
    =&2\frac{\partial}{\partial\omega}v^{\top}\nabla_{x}\log p_{\theta}(x)+\sum_{|\bm{\alpha}|=3}\left(R_{\bm{\alpha}}^{\omega}(x+v)-R_{\bm{\alpha}}^{\omega}(x-v)\right)\cdot v^{\bm{\alpha}}\text{.}
\end{split}
\end{equation}
Similar for the expansion of $\log p_{\theta}(x+v)$, the remainder is
\begin{equation}
    R_{\bm{\alpha}}(x+v)=\frac{|\bm{\alpha}|}{\bm{\alpha}!}\int_{0}^{1}(1-t)^{|\bm{\alpha}|-1}D^{\bm{\alpha}}\log p_{\theta}(x+t\cdot v)dt
\end{equation}
and we can obtain the uniform upper bound on the compact set $\overline{B}_{\epsilon_{0}}$ as
\begin{equation}
    |R_{\bm{\alpha}}(x+v)|\leq U^{+}(\bm{\alpha})\leq\max_{\bm{\alpha}}U^{+}(\bm{\alpha})=U^{+}\text{.}
\end{equation}
We denote the bound for $R_{\bm{\alpha}}(x-v)$ as $U^{-}$ and further have
\begin{equation}
    \begin{split}
    &\log p_{\theta}(x+v)-\log p_{\theta}(x-v)\\
    =&2v^{\top}\nabla_{x}\log p_{\theta}(x)+\sum_{|\bm{\alpha}|=3}\left(R_{\bm{\alpha}}(x+v)-R_{\bm{\alpha}}(x-v)\right)\cdot v^{\bm{\alpha}}\text{.}
\end{split}
\end{equation}
We denote $\Delta R_{\bm{\alpha}}=R_{\bm{\alpha}}(x+v)-R_{\bm{\alpha}}(x-v)$ and $\Delta R_{\bm{\alpha}}^{\omega,+}=R_{\bm{\alpha}}^{\omega}(x+v)+R_{\bm{\alpha}}^{\omega}(x-v)$ and $\Delta R_{\bm{\alpha}}^{\omega,-}=R_{\bm{\alpha}}^{\omega}(x+v)-R_{\bm{\alpha}}^{\omega}(x-v)$ for notation compactness. Thus for $\forall (x,\theta)\in B$ and $\|v\|_{2}=\epsilon$, we obtain the partial derivative of $\mathcal{J}_{\text{FD-SSM}}(x,v;\theta)$ as
\begin{equation*}
\begin{split}
    \!\!\!&\frac{\partial}{\partial \omega}\mathcal{J}_{\text{FD-SSM}}(x,v;\theta)\\
    =&\frac{1}{\epsilon^2}\left(v^{\top}\nabla_{x}\log p_{\theta}(x)+\frac{1}{2}\sum_{|\bm{\alpha}|=3}\Delta R_{\bm{\alpha}}\cdot v^{\bm{\alpha}}\right)\cdot\left(\frac{\partial}{\partial\omega}v^{\top}\nabla_{x}\log p_{\theta}(x)+\frac{1}{2}\sum_{|\bm{\alpha}|=3}\Delta R_{\bm{\alpha}}^{\omega,-}\cdot v^{\bm{\alpha}}\right)\\
    &+\frac{1}{\epsilon^2}\left(\frac{\partial}{\partial\omega}v^{\top}\nabla_{x}^{2}\log p_{\theta}(x)v+\sum_{|\bm{\alpha}|=3}\Delta R_{\bm{\alpha}}^{\omega,+}\cdot v^{\bm{\alpha}}\right)\\
    =&\frac{1}{\epsilon^2}\left(v^{\top}\nabla_{x}\log p_{\theta}(x)\cdot\frac{\partial}{\partial\omega}v^{\top}\nabla_{x}\log p_{\theta}(x)+\frac{\partial}{\partial\omega}v^{\top}\nabla_{x}^{2}\log p_{\theta}(x)v\right)\\
    &+\frac{1}{\epsilon^2}\left(v^{\top}\nabla_{x}\log p_{\theta}(x)\cdot\frac{1}{2}\sum_{|\bm{\alpha}|=3}\Delta R_{\bm{\alpha}}^{\omega,-}\cdot v^{\bm{\alpha}}+\frac{\partial}{\partial\omega}v^{\top}\nabla_{x}\log p_{\theta}(x)\cdot\frac{1}{2}\sum_{|\bm{\alpha}|=3}\Delta R_{\bm{\alpha}}\cdot v^{\bm{\alpha}}\right)\\
    &+\frac{1}{\epsilon^2}\left(\left(\frac{1}{2}\sum_{|\bm{\alpha}|=3}\Delta R_{\bm{\alpha}}\cdot v^{\bm{\alpha}}\right)\cdot\left(\frac{1}{2}\sum_{|\bm{\alpha}|=3}\Delta R_{\bm{\alpha}}^{\omega,-}\cdot v^{\bm{\alpha}}\right)+\sum_{|\bm{\alpha}|=3}\Delta R_{\bm{\alpha}}^{\omega,+}\cdot v^{\bm{\alpha}}\right)\text{.}
\end{split}
\end{equation*}
Note that the first term in the above equals to $\frac{\partial}{\partial \omega}\mathcal{J}_{\text{SSM}}(x,v;\theta)$. Due to the continuity of the norm functions $\|\nabla_{x}\log p_{\theta}(x)\|_{2}$ and $\|\nabla_{x}\frac{\partial}{\partial\omega}\log p_{\theta}(x)\|_{2}$ on the compact set $\overline{B}_{\epsilon_{0}}$, we denote their upper bound as $G$ and $G_{\omega}$, respectively. Then we have $|v^{\top}\nabla_{x}\log p_{\theta}(x)|\leq \epsilon G$ and $\frac{\partial}{\partial\omega}v^{\top}\nabla_{x}\log p_{\theta}(x)=v^{\top}\nabla_{x}\frac{\partial}{\partial\omega}\log p_{\theta}(x)\leq \epsilon G_{\omega}$. Now we can derive the bound between the partial derivatives of FD-SSM and SSM as
\begin{equation}
\begin{split}
    &\left|\frac{\partial}{\partial \omega}\mathcal{J}_{\text{FD-SSM}}(x,v;\theta)-\frac{\partial}{\partial \omega}\mathcal{J}_{\text{SSM}}(x,v;\theta)\right|\\
    <&\frac{1}{2\epsilon^2}\left(\epsilon G  d^{3}\epsilon^3\Delta U_{\omega}+\epsilon G_{\omega} d^{3}\epsilon^3\Delta U+\frac{1}{2}\Delta U_{\omega}\Delta U d^{6}\epsilon^{6}+2\Delta U_{\omega}d^{3}\epsilon^3\right)\\
    <&\epsilon\cdot\frac{1}{2}\left(G  d^{2}\Delta U_{\omega}+G_{\omega}  d^{2}\Delta U+\frac{1}{2}\Delta U_{\omega}\Delta U d^{3}+2\Delta U_{\omega}d^{3}\right)\text{, holds when }\epsilon<\frac{1}{d}\text{,}
\end{split}
\end{equation}
where we denote $\Delta U = U^{+}+U^{-}$ and $\Delta U_{\omega} = U_{\omega}^{+}+U_{\omega}^{-}$. By setting $\epsilon_{0}<\frac{1}{d}$, we can omit the condition $\epsilon<\frac{1}{d}$ since $\epsilon<\epsilon_{0}=\min(\epsilon_{0},\frac{1}{d})$. Note that the condition $\epsilon<\frac{1}{d}$ can be generalize to, e.g., $\epsilon<1$ without changing our conclusions. Then it is easy to show that
\begin{equation}
\begin{split}
    &\left\|\nabla_{\theta}\mathcal{J}_{\text{FD-SSM}}(x,v;\theta)-\nabla_{\theta}\mathcal{J}_{\text{SSM}}(x,v;\theta)\right\|_{2}\\
    \leq&\text{dim}(\mathcal{S})\cdot\max_{\omega\in\theta}\left|\frac{\partial}{\partial \omega}\mathcal{J}_{\text{FD-SSM}}(x,v;\theta)-\frac{\partial}{\partial \omega}\mathcal{J}_{\text{SSM}}(x,v;\theta)\right|\\
    <&\epsilon\cdot\text{dim}(\mathcal{S})\cdot\max_{\omega\in\theta}M_{\omega}\text{,}\\
    &\text{where }M_{\omega}=\frac{1}{2}\left(G  d^{2}\Delta U_{\omega}+G_{\omega}  d^{2}\Delta U+\frac{1}{2}\Delta U_{\omega}\Delta U d^{3}+2\Delta U_{\omega}d^{3}\right)\text{.}
\end{split}
\end{equation}
Just to emphasize here, the bound above uniformly holds for $\forall (x,\theta)\in B$ and $v\in\R^{d}, \|v\|_{2}=\epsilon<\epsilon_{0}$. We have the simple fact that give two vectors $a$ and $b$, if there is $\|a-b\|_{2}<\|b\|_{2}$, then their angle is $\angle(a,b)\leq \arcsin(\|a-b\|_{2}/\|b\|_{2})$. So finally we can derive the angle
\begin{equation}
\begin{split}
    &\angle\left(\nabla_{\theta}\mathcal{J}_{\text{FD-SSM}}(x,v;\theta),\nabla_{\theta}\mathcal{J}_{\text{SSM}}(x,v;\theta)\right)\\
    \leq&\arcsin\left(\frac{\left\|\nabla_{\theta}\mathcal{J}_{\text{FD-SSM}}(x,v;\theta)-\nabla_{\theta}\mathcal{J}_{\text{SSM}}(x,v;\theta)\right\|_{2}}{\left\|\nabla_{\theta}\mathcal{J}_{\text{SSM}}(x,v;\theta)\right\|_{2}}\right)\\
    \leq&\arcsin\left(\frac{\epsilon\cdot\text{dim}(\mathcal{S})\cdot\max_{\omega\in\theta}M_{\omega}}{\min_{(x,\theta)\in B, \|v\|_{2}<\epsilon_{0}}\left\|\nabla_{\theta}\mathcal{J}_{\text{SSM}}(x,v;\theta)\right\|_{2}}\right)\\
    <&\eta\text{,}
\end{split}
\end{equation}
where $\min_{(x,\theta)\in B, \|v\|_{2}<\epsilon_{0}}\left\|\nabla_{\theta}\mathcal{J}_{\text{SSM}}(x,v;\theta)\right\|_{2}$ must exist and larger than $0$ due to the continuity of $\nabla_{\theta}\mathcal{J}_{\text{SSM}}(x,v;\theta)$ and the condition that $\|\nabla_{\theta}\mathcal{J}_{\text{SSM}}(x,v;\theta)\|_{2}>0$ on the compact set. So we only need to choose $\xi$ as
\begin{equation}
    \xi=\frac{\sin\eta \cdot\min_{(x,\theta)\in B, \|v\|_{2}<\epsilon_{0}}\left\|\nabla_{\theta}\mathcal{J}_{\text{SSM}}(x,v;\theta)\right\|_{2}}{\text{dim}(\mathcal{S})\cdot\max_{\omega\in\theta}M_{\omega}}\text{.}
\end{equation}
When we choose $\|v\|_{2}=\epsilon<\min(\epsilon_{0},\xi)$, we can guarantee the angle between $\nabla_{\theta}\mathcal{J}_{\text{FD-SSM}}(x,v;\theta)$ and $\nabla_{\theta}\mathcal{J}_{\text{SSM}}(x,v;\theta)$ to be uniformly less than $\eta$ on $B$.
\qed

\subsection{Proof of Theorem 2}
We consider in the compact set $\overline{B}_{\epsilon_{0}}$ defined in Lemma 2. The assumptions for general stochastic optimization include:

{
\renewcommand{\labelitemi}{$\vcenter{\hbox{\tiny$\bullet$}}$}
\begin{itemize}
    \item \textbf{(\romannumeral 1)} The condition of Corollary 4.12 in~\citet{bottou2018optimization}: $\mathcal{J}_{\text{FD-SSM}}(\theta)$ is twice-differentiable with $\theta$;
    \item \textbf{(\romannumeral 2)} The Assumption 4.1 in~\citet{bottou2018optimization}: the gradient $\nabla_{\theta}\mathcal{J}_{\text{FD-SSM}}(\theta)$ is Lipschitz;
    \item \textbf{(\romannumeral 3)} The Assumption 4.3 in~\citet{bottou2018optimization}: the first and second moments of the stochastic gradients are bounded by the expected gradients;
    \item \textbf{(\romannumeral 4)} The stochastic step size $\alpha_{k}$ satisfies the diminishing condition in~\citet{bottou2018optimization}: $\sum_{k=1}^{\infty}\alpha_{k}=\infty$, $\sum_{k=1}^{\infty}\alpha_{k}^{2}<\infty$;
    \item \textbf{(\romannumeral 5)} The condition of Lemma 2 holds in each step $k$ of stochastic gradient update.
\end{itemize}
}

Note that the condition \textbf{(\romannumeral 5)} only holds in the compact set $\overline{B}_{\epsilon_{0}}$, but we can choose it to be large enough to contain $(x,\theta_{k}), x\sim p(x)$, as well as containing the neighborhood of stationary points of $\nabla_{\theta}\mathcal{J}_{\text{SSM}}(\theta)$. These can be achieved by setting $\epsilon\rightarrow 0$. Thus we have
\begin{equation}
   \lim_{k\rightarrow\infty,\epsilon\rightarrow 0}\mathbb{E}\left[\left\|\nabla_{\theta}\mathcal{J}_{\text{SSM}}(\theta_{k})\right\|_{2}\right]=0\text{.}
\end{equation}
This means that stochastically optimizing the FD-SSM objective can make the parameters $\theta$ converge to the stationary point of the SSM objective when $\epsilon\rightarrow 0$.
\qed

\section{Extended conclusions}
In this section we provide extended and supplementary conclusions for the main text.

\subsection{Parallel computing on dependent operations}
For the dependent operations like those in the gradient-based SM methods, it is possible to execute them on different devices via asynchronous parallelism~\citep{lian2015asynchronous}. However, this asynchronous parallelization needs to perform across different data batches, requires complex design on the synchronization mechanism, and could introduce extra bias when updating the model parameters. These difficulties usually outweigh the gain from paralleling the operations in the gradient-based SM methods. In contrast, for our FD-based SM methods, the decomposed independent operations can be easily executed in a synchronous manner, which is further compatible with data or model parallelism.

\subsection{Scaling the projection vector in training objectives}
Below we explain why the scale of the random projection $v$ will not affect the training of SM objectives.
For the original SM objective, we have
\begin{equation}
\begin{split}
    \mathcal{J}_{\text{SM}}(\theta)=&\mathbb{E}_{p_{\textup{data}}(x)}\left[\text{tr}(\nabla_{x}^{2}\log p_{\theta}(x))+\frac{1}{2}\|\nabla_{x}\log p_{\theta}(x)\|_{2}^{2}\right]\\
    =&\mathbb{E}_{p_{\textup{data}}(x)}\left[\sum_{i=1}^{d}\bm{e}_{i}^{\top}\nabla_{x}^{2}\log p_{\theta}(x)\bm{e}_{i}+\frac{1}{2}\sum_{i=1}^{d}\left(\bm{e}_{i}^{\top}\nabla_{x}\log p_{\theta}(x)\right)^{2}\right]\textup{.}
\end{split}
\end{equation}
When we scale the basis vector $\bm{e}_{i}$ with a small value ${\epsilon'}$, i.e., $\bm{e}_{i}\rightarrow{\epsilon'}\bm{e}_{i}$, we have
\begin{equation}
\begin{split}
   &\mathbb{E}_{p_{\textup{data}}(x)}\left[\sum_{i=1}^{d}({\epsilon'}\bm{e}_{i})^{\top}\nabla_{x}^{2}\log p_{\theta}(x)({\epsilon'}\bm{e}_{i})+\frac{1}{2}\sum_{i=1}^{d}\left(({\epsilon'}\bm{e}_{i})^{\top}\nabla_{x}\log p_{\theta}(x)\right)^{2}\right]\\
   =&{\epsilon'}^{2}\mathbb{E}_{p_{\textup{data}}(x)}\left[\sum_{i=1}^{d}\bm{e}_{i}^{\top}\nabla_{x}^{2}\log p_{\theta}(x)\bm{e}_{i}+\frac{1}{2}\sum_{i=1}^{d}\left(\bm{e}_{i}^{\top}\nabla_{x}\log p_{\theta}(x)\right)^{2}\right]
   ={\epsilon'}^{2}\mathcal{J}_{\text{SM}}(\theta)\textup{.}
\end{split}
\end{equation}
Thus we can simply divide the objective by ${\epsilon'}^{2}$ to recover the original SM objective $\mathcal{J}_{\text{SM}}(\theta)$. Similarly, for the DSM objective we have
\begin{equation}
\begin{split}
    \mathcal{J}_{\text{DSM}}(\theta) =& \frac{1}{d}\mathbb{E}_{p_{\textup{data}}(x)}\mathbb{E}_{p_{\sigma}(\widetilde{x}|x)} \left[\left\|\nabla_{\widetilde{x}} \log p_{\theta}(\widetilde{x}) + \frac{\widetilde{x}-x}{\sigma^2}\right\|_2^2\right]\\
    =&\frac{1}{d}\mathbb{E}_{p_{\textup{data}}(x)}\mathbb{E}_{p_{\sigma}(\widetilde{x}|x)} \left[\sum_{i=1}^{d}\left(\bm{e}_{i}^{\top}\nabla_{\widetilde{x}} \log p_{\theta}(\widetilde{x}) + \frac{\bm{e}_{i}^{\top}(\widetilde{x}-x)}{\sigma^2}\right)^2\right]\text{.}
\end{split}
\end{equation}
When we scale the basis vector $\bm{e}_{i}$ with a small value ${\epsilon'}$, i.e., $\bm{e}_{i}\rightarrow{\epsilon'}\bm{e}_{i}$, we also have
\begin{equation}
\begin{split}
    &\frac{1}{d}\mathbb{E}_{p_{\textup{data}}(x)}\mathbb{E}_{p_{\sigma}(\widetilde{x}|x)} \left[\sum_{i=1}^{d}\left(({\epsilon'}\bm{e}_{i})^{\top}\nabla_{\widetilde{x}} \log p_{\theta}(\widetilde{x}) + \frac{({\epsilon'}\bm{e}_{i})^{\top}(\widetilde{x}-x)}{\sigma^2}\right)^2\right]\\
    =&\frac{{\epsilon'}^{2}}{d}\mathbb{E}_{p_{\textup{data}}(x)}\mathbb{E}_{p_{\sigma}(\widetilde{x}|x)} \left[\sum_{i=1}^{d}\left(\bm{e}_{i}^{\top}\nabla_{\widetilde{x}} \log p_{\theta}(\widetilde{x}) + \frac{\bm{e}_{i}^{\top}(\widetilde{x}-x)}{\sigma^2}\right)^2\right]={\epsilon'}^{2}\mathcal{J}_{\text{DSM}}(\theta)\text{.}
\end{split}
\end{equation}
Thus we can divide by ${\epsilon'}^{2}$ to recover the DSM objective $\mathcal{J}_{\text{DSM}}(\theta)$. Finally as to SSM, we have
\begin{equation}
    \mathcal{J}_{\text{SSM}}(\theta)=\frac{1}{C_{v}}\mathbb{E}_{p_{\textup{data}}(x)}\mathbb{E}_{p_{v}(v)}\left[v^{\top}\nabla^{2}_{x}\log p_{\theta}(x)v+\frac{1}{2}\left(v^{\top}\nabla_{x}\log p_{\theta}(x)\right)^{2}\right]\text{.}
\end{equation}
When we scale the random projection $v$ with a small value ${\epsilon'}$, i.e., $v\rightarrow{\epsilon'}v$, we should not that the adaptive factor $C_{v}$ will also be scaled to ${\epsilon'}^{2}C_{v}$, then we can derive
\begin{equation}
    \frac{1}{{\epsilon'}^{2}C_{v}}\mathbb{E}_{p_{\textup{data}}(x)}\mathbb{E}_{p_{v}(v)}\left[({\epsilon'}v)^{\top}\nabla^{2}_{x}\log p_{\theta}(x)({\epsilon'}v)+\frac{1}{2}\left(({\epsilon'}v)^{\top}\nabla_{x}\log p_{\theta}(x)\right)^{2}\right]=\mathcal{J}_{\text{SSM}}(\theta)\text{.}
\end{equation}
This indicates that the SSM objective is already invariant to the scaling of $v$. It is trivial to also divide similar factors as $C_{v}$ in SM and DSM to result in similarly invariant objectives.

\subsection{Mild regularity conditions for the FD-based SM methods}
The mild conditions for the original gradient-based SM methods~\citep{hyvarinen2005estimation,song2019sliced} include: (\textbf{\romannumeral 1}) $p_{\text{data}}(x)$ and $p_{\theta}(x)$ are both twice-differentiable on $\R^{d}$; (\textbf{\romannumeral 2}) $\mathbb{E}_{p_{\text{data}}(x)}[\|\nabla_{x}\log p_{\theta}(x)\|_{2}^{2}]$ and $\mathbb{E}_{p_{\text{data}}(x)}[\|\nabla_{x}\log p_{\text{data}}(x)\|_{2}^{2}]$ are finite for any $\theta$; (\textbf{\romannumeral 3}) There is $\lim_{\|x\|\rightarrow \infty}p_{\textup{data}}(x)\nabla_{x}\log p_{\theta}(x)=0$ holds for any $\theta$. Here we provide two additional regularity conditions which are sufficient to guarantee the $o(\epsilon)$ or $\mathcal{O}(\epsilon^2)$ approximation error of FD-SSM and FD-DSM: (\textbf{\romannumeral 4}) $p_{\theta}(x)$ is four-times continuously differentiable on $\R^{d}$; (\textbf{\romannumeral 5}) There is $\mathbb{E}_{p_{\textup{data}}(x)}[|D^{\bm{\alpha}}\log p_{\theta}(x)|]<\infty$ holds for any $\theta$ and $|\bm{\alpha}|=4$, where $D$ and $\bm{\alpha}$ are defined in Eq.~(\ref{Ddefinition}). The proof is almost the same as it for Theorem {\color{red} 1} under Lagrange's remainder. 

\textbf{Remark.} Note that the condition (\textbf{\romannumeral 4}) holds when we apply, e.g., average pooling layers and Softplus activation in the neural network models, while the condition (\textbf{\romannumeral 5}) always holds as long as the support set of $p_{\textup{data}}(x)$ is bounded, e.g., for RGB-based image tasks there is $x\in[0,255]^{d}$.

\subsection{DSM under sliced Wasserstein distance}
To construct the FD instantiation for DSM, we first cast the original objective of DSM into sliced Wasserstein distance~\citep{rabin2011wasserstein} with random projection $v$. Since there is $\mathbb{E}_{p_{\epsilon}(v)}\left[vv^{\top}\right]=\frac{\epsilon^2I}{d}$, we can rewrite the objective of DSM with Gaussian noise distribution as
\begin{equation}
    \mathcal{J}_{\text{DSM}}(\theta) = \frac{1}{\epsilon^2} \mathbb{E}_{p_{\textup{data}}(x)}\mathbb{E}_{p_{\sigma}(\widetilde{x}|x)}\mathbb{E}_{p_{\epsilon}(v)} \left[\left(v^{\top}\nabla_{\widetilde{x}} \log p_{\theta}(\widetilde{x}) + \frac{v^{\top}(\widetilde{x}-x)}{\sigma^2}\right)^2\right]\text{.}
    \label{DSM_trace}
\end{equation}
In this case, there is $\frac{v^{\top}(\widetilde{x}-x)}{\sigma^2}=\mathcal{O}(\epsilon)$ with high probability, thus we can approximate $v^{\top}\nabla_{\widetilde{x}} \log p_{\theta}(\widetilde{x})$ according to our FD decomposition.

\subsection{Consistency between DSM and FD-DSM}
\begin{theoremappendix}
\label{theorem2}
Let $\mathcal{S}$ be the parameter space of $\theta$, $B$ be a bounded set in the space of $\R^{d}\times\mathcal{S}$, and $B_{\epsilon_{0}}$ is the $\epsilon_{0}$-neighbourhood of $B$ for certain $\epsilon_{0}>0$. Then under the condition that $\log p_{\theta}(\widetilde{x})$ is three times continuously differentiable w.r.t. $(\widetilde{x},\theta)$ and $\|\nabla_{\theta}\mathcal{J}_{\textup{DSM}}(x,\widetilde{x},v;\theta)\|_{2}>0$ in the closure of $B_{\epsilon_{0}}$, we have $\forall \eta>0$, $\exists \xi>0$, such that
\begin{equation}
    \angle\left(\nabla_{\theta}\mathcal{J}_{\textup{FD-DSM}}(x,\widetilde{x},v;\theta), \nabla_{\theta}\mathcal{J}_{\textup{DSM}}(x,\widetilde{x},v;\theta)\right)<\eta
\end{equation}
uniformly holds for $\forall(\widetilde{x},\theta)\in B, v\in\R^{d}, \|v\|_{2}=\epsilon<\min(\xi,\epsilon_{0})$ and $x$ in any bounded subset of $\R^{d}$. Here $\angle(\cdot,\cdot)$ denotes the angle between two vectors. The arguments $x,\widetilde{x},v$ in the objectives indicate the losses at that point.
\end{theoremappendix}
\emph{Proof.} Following the routines and notations in the proof of Lemma 2, we investigate the gradient $\nabla_{\theta}\mathcal{J}_{\text{FD-DSM}}(x,\widetilde{x},v;\theta)$, whose elements consist of $\frac{\partial}{\partial \omega}\mathcal{J}_{\text{FD-DSM}}(x,\widetilde{x},v;\theta)$ for $\omega\in\theta$. When $\log p_{\theta}(\widetilde{x})$ is three-times-differentiable in $\overline{B}_{\epsilon_{0}}$, we can obtain
\begin{equation}
\begin{split}
    &\frac{\partial}{\partial\omega}\log p_{\theta}(\widetilde{x}+v)\\
    =& \frac{\partial}{\partial\omega}\log p_{\theta}(\widetilde{x})+v^{\top}\nabla_{\widetilde{x}}\frac{\partial}{\partial\omega}\log p_{\theta}(\widetilde{x})+\frac{1}{2}v^{\top}\nabla_{\widetilde{x}}^{2}\frac{\partial}{\partial\omega}\log p_{\theta}(\widetilde{x})v+\sum_{|\bm{\alpha}|=3}R_{\bm{\alpha}}^{\omega}(\widetilde{x}+v)\cdot v^{\bm{\alpha}}\\
    =&\frac{\partial}{\partial\omega}\left[\log p_{\theta}(\widetilde{x})+v^{\top}\nabla_{\widetilde{x}}\log p_{\theta}(\widetilde{x})+\frac{1}{2}v^{\top}\nabla_{\widetilde{x}}^{2}\log p_{\theta}(\widetilde{x})v\right]+\sum_{|\bm{\alpha}|=3}R_{\bm{\alpha}}^{\omega}(\widetilde{x}+v)\cdot v^{\bm{\alpha}}\text{,}\\
    &\text{where }R_{\bm{\alpha}}^{\omega}(\widetilde{x}+v)=\frac{|\bm{\alpha}|}{\bm{\alpha}!}\int_{0}^{1}(1-t)^{|\bm{\alpha}|-1}D^{\bm{\alpha}}\frac{\partial}{\partial\omega}\log p_{\theta}(\widetilde{x}+t\cdot v)dt\text{,}
    \label{appendixequ2}
\end{split}
\end{equation}
Then we can further obtain that
\begin{equation*}
   \! \frac{\partial}{\partial\omega}\left[\log p_{\theta}(\widetilde{x}\!+\!v)\!-\!\log p_{\theta}(\widetilde{x}\!-\!v)\right]=2\frac{\partial}{\partial\omega}v^{\top}\nabla_{\widetilde{x}}\log p_{\theta}(\widetilde{x})\!+\!\sum_{|\bm{\alpha}|=3}\left(R_{\bm{\alpha}}^{\omega}(\widetilde{x}\!+\!v)\!-\!R_{\bm{\alpha}}^{\omega}(\widetilde{x}\!-\!v)\right)\cdot v^{\bm{\alpha}}\text{.}
\end{equation*}
Due to the continuity of $R_{\bm{\alpha}}^{\omega}(\widetilde{x}+v)$ and $R_{\bm{\alpha}}^{\omega}(\widetilde{x}-v)$ on the compact set $\overline{B}_{\epsilon_{0}}$, they have the uniform absolute upper bounds $U^{+}_{\omega}$ and $U^{-}_{\omega}$, respectively. Similarly, we have
\begin{equation*}
   \! \log p_{\theta}(\widetilde{x}\!+\!v)\!-\!\log p_{\theta}(\widetilde{x}\!-\!v)=2v^{\top}\nabla_{\widetilde{x}}\log p_{\theta}(\widetilde{x})\!+\!\sum_{|\bm{\alpha}|=3}\left(R_{\bm{\alpha}}(\widetilde{x}\!+\!v)\!-\!R_{\bm{\alpha}}(\widetilde{x}\!-\!v)\right)\cdot v^{\bm{\alpha}}\text{,}
\end{equation*}
where the uniform absolute upper bounds for $R_{\bm{\alpha}}(\widetilde{x}+v)$ and $R_{\bm{\alpha}}(\widetilde{x}-v)$ are $U^{+}$ and $U^{-}$, respectively. Besides, note that the terms $\frac{x-\widetilde{x}}{\sigma}$ in the DSM / FD-DSM objectives are independent of $\theta$. We denote $\Delta R_{\bm{\alpha}}=R_{\bm{\alpha}}(\widetilde{x}+v)-R_{\bm{\alpha}}(\widetilde{x}-v)$ and $\Delta R_{\bm{\alpha}}^{\omega}=R_{\bm{\alpha}}^{\omega}(\widetilde{x}+v)-R_{\bm{\alpha}}^{\omega}(\widetilde{x}-v)$ for notation compactness. Thus for $\forall (\widetilde{x},\theta)\in B$ and $\|v\|_{2}=\epsilon, x\in\R^{d}$, we can obtain the partial derivative of $\mathcal{J}_{\text{FD-DSM}}(x,\widetilde{x},v;\theta)$ as
\begin{equation*}
\begin{split}
   &\frac{\partial}{\partial \omega}\mathcal{J}_{\text{FD-DSM}}(x,\widetilde{x},v;\theta)\\
    =&\frac{1}{2\epsilon^2}\left(2v^{\top}\nabla_{\widetilde{x}}\log p_{\theta}(\widetilde{x})+\sum_{|\bm{\alpha}|=3}\Delta R_{\bm{\alpha}}\cdot v^{\bm{\alpha}}+\frac{2v^{\top}(\widetilde{x}\!-\!x)}{\sigma^2}\right)\cdot\left(2\frac{\partial}{\partial\omega}v^{\top}\nabla_{\widetilde{x}}\log p_{\theta}(\widetilde{x})+\sum_{|\bm{\alpha}|=3}\Delta R_{\bm{\alpha}}^{\omega}\cdot v^{\bm{\alpha}}\right)\\
    =&\frac{1}{\epsilon^2}\frac{\partial}{\partial\omega}\left(v^{\top}\nabla_{\widetilde{x}}\log p_{\theta}(\widetilde{x})+\frac{v^{\top}(\widetilde{x}\!-\!x)}{\sigma^2}\right)^2+\frac{1}{2\epsilon^2}\left(\sum_{|\bm{\alpha}|=3}\Delta R_{\bm{\alpha}}^{\omega}\cdot v^{\bm{\alpha}}\right)\cdot\left(\sum_{|\bm{\alpha}|=3}\Delta R_{\bm{\alpha}}\cdot v^{\bm{\alpha}}\right)\\
    &+\frac{1}{\epsilon^2}\left(\left(v^{\top}\nabla_{\widetilde{x}}\log p_{\theta}(\widetilde{x})+\frac{v^{\top}(\widetilde{x}\!-\!x)}{\sigma^2}\right)\sum_{|\bm{\alpha}|=3}\Delta R_{\bm{\alpha}}^{\omega}\cdot v^{\bm{\alpha}}+\frac{\partial}{\partial\omega}v^{\top}\nabla_{\widetilde{x}}\log p_{\theta}(\widetilde{x})\sum_{|\bm{\alpha}|=3}\Delta R_{\bm{\alpha}}\cdot v^{\bm{\alpha}}\right)\text{,}
\end{split}
\end{equation*}
where the first term equals to $\frac{\partial}{\partial \omega}\mathcal{J}_{\text{DSM}}(x,\widetilde{x},v;\theta)$. Due to the continuity of the norm functions $\|\nabla_{\widetilde{x}}\log p_{\theta}(\widetilde{x})\|_{2}$ and $\|\nabla_{\widetilde{x}}\frac{\partial}{\partial\omega}\log p_{\theta}(\widetilde{x})\|_{2}$ on the compact set $\overline{B}_{\epsilon_{0}}$, we denote their upper bound as $G$ and $G_{\omega}$, respectively. Then we have $|v^{\top}\nabla_{\widetilde{x}}\log p_{\theta}(\widetilde{x})|\leq \epsilon G$ and $\frac{\partial}{\partial\omega}v^{\top}\nabla_{\widetilde{x}}\log p_{\theta}(\widetilde{x})=v^{\top}\nabla_{\widetilde{x}}\frac{\partial}{\partial\omega}\log p_{\theta}(\widetilde{x})\leq \epsilon G_{\omega}$. Besides, since $\widetilde{x}$ and $x$ both come from bounded sets, we have an upper bound of $v^{\top}(\widetilde{x}-x)\leq\epsilon \sigma^2 G_{x}$. Now we can derive the bound between the partial derivatives of FD-DSM and DSM as
\begin{equation}
\begin{split}
    &|\frac{\partial}{\partial \omega}\mathcal{J}_{\text{FD-DSM}}(x,\widetilde{x},v;\theta)-\frac{\partial}{\partial \omega}\mathcal{J}_{\text{DSM}}(x,\widetilde{x},v;\theta)|\\
    <&\frac{1}{\epsilon^2}\left(\frac{1}{2}\Delta U\Delta U_{\omega}d^{6}\epsilon^{6}+(G+G_{x})\Delta U_{\omega}d^{3}\epsilon^{3}+G_{\omega}\Delta U d^{3}\epsilon^{3}\right)\\
    <&\epsilon\cdot\left(\frac{1}{2}\Delta U\Delta U_{\omega}d^{3}+(G+G_{x})\Delta U_{\omega}d^{3}+G_{\omega}\Delta U d^{3}\right)\text{, holds when }\epsilon<\frac{1}{d}\text{,}
\end{split}
\end{equation}
where we denote $\Delta U = U^{+}+U^{-}$ and $\Delta U_{\omega} = U_{\omega}^{+}+U_{\omega}^{-}$. By setting $\epsilon_{0}<\frac{1}{d}$, we can omit the condition $\epsilon<\frac{1}{d}$ since $\epsilon<\epsilon_{0}=\min(\epsilon_{0},\frac{1}{d})$. Note that the condition $\epsilon<\frac{1}{d}$ can be generalize to, e.g., $\epsilon<1$ without changing our conclusions. Then it is easy to show that
\begin{equation}
\begin{split}
    &\left\|\nabla_{\theta}\mathcal{J}_{\text{FD-DSM}}(x,\widetilde{x},v;\theta)-\nabla_{\theta}\mathcal{J}_{\text{DSM}}(x,\widetilde{x},v;\theta)\right\|_{2}\\
    \leq&\text{dim}(\mathcal{S})\cdot\max_{\omega\in\theta}\left|\frac{\partial}{\partial \omega}\mathcal{J}_{\text{FD-DSM}}(x,\widetilde{x},v;\theta)-\frac{\partial}{\partial \omega}\mathcal{J}_{\text{DSM}}(x,\widetilde{x},v;\theta)\right|\\
    <&\epsilon\cdot\text{dim}(\mathcal{S})\cdot\max_{\omega\in\theta}M_{\omega}\text{,}\\
    &\text{where }M_{\omega}=\frac{1}{2}\Delta U\Delta U_{\omega}d^{3}+(G+G_{x})\Delta U_{\omega}d^{3}+G_{\omega}\Delta U d^{3}\text{.}
\end{split}
\end{equation}
Just to emphasize here, the bound above uniformly holds for $\forall (\widetilde{x},\theta)\in B$ and $v\in\R^{d}, \|v\|_{2}=\epsilon<\epsilon_{0}$ and $x$ from any bounded set ($x$ is inherently bounded when we consider, e.g., pixel input space). So finally we can derive the angle
\begin{equation}
\begin{split}
    &\angle\left(\nabla_{\theta}\mathcal{J}_{\text{FD-DSM}}(x,\widetilde{x},v;\theta),\nabla_{\theta}\mathcal{J}_{\text{DSM}}(x,\widetilde{x},v;\theta)\right)\\
    \leq&\arcsin\left(\frac{\left\|\nabla_{\theta}\mathcal{J}_{\text{FD-DSM}}(x,\widetilde{x},v;\theta)-\nabla_{\theta}\mathcal{J}_{\text{DSM}}(x,\widetilde{x},v;\theta)\right\|_{2}}{\left\|\nabla_{\theta}\mathcal{J}_{\text{DSM}}(x,\widetilde{x},v;\theta)\right\|_{2}}\right)\\
    \leq&\arcsin\left(\frac{\epsilon\cdot\text{dim}(\mathcal{S})\cdot\max_{\omega\in\theta}M_{\omega}}{\min_{(\widetilde{x},\theta)\in B, \|v\|_{2}<\epsilon_{0}}\left\|\nabla_{\theta}\mathcal{J}_{\text{DSM}}(x,\widetilde{x},v;\theta)\right\|_{2}}\right)\\
    <&\eta\text{,}
\end{split}
\end{equation}
where $\min_{(\widetilde{x},\theta)\in B, \|v\|_{2}<\epsilon_{0}}\left\|\nabla_{\theta}\mathcal{J}_{\text{DSM}}(x,\widetilde{x},v;\theta)\right\|_{2}$ must exist and larger than $0$ due to the continuity of $\nabla_{\theta}\mathcal{J}_{\text{DSM}}(x,\widetilde{x},v;\theta)$ and the condition that $\|\nabla_{\theta}\mathcal{J}_{\text{DSM}}(x,\widetilde{x},v;\theta)\|_{2}>0$ on the compact set. So we only need to choose $\xi$ as
\begin{equation}
    \xi=\frac{\sin\eta \cdot\min_{(\widetilde{x},\theta)\in B, \|v\|_{2}<\epsilon_{0}}\left\|\nabla_{\theta}\mathcal{J}_{\text{DSM}}(x,\widetilde{x},v;\theta)\right\|_{2}}{\text{dim}(\mathcal{S})\cdot\max_{\omega\in\theta}M_{\omega}}\text{.}
\end{equation}
When we choose $\|v\|_{2}=\epsilon<\min(\epsilon_{0},\xi)$, we can guarantee the angle between $\nabla_{\theta}\mathcal{J}_{\text{FD-DSM}}(x,\widetilde{x},v;\theta)$ and $\nabla_{\theta}\mathcal{J}_{\text{DSM}}(x,\widetilde{x},v;\theta)$ to be uniformly less than $\eta$ on $B$.
\qed

\subsection{Application on the latent variable models}
For the latent variable models (LVMs), the log-likelihood is usually intractable. Unlike EBMs, this intractability cannot be easily eliminated by taking gradients. Recently, the proposed SUMO~\citep{Luo2020SUMO} can provide an unbiased estimator for the intractable $\log p_{\theta}(x)$, which is defined as
\begin{equation}
    \text{SUMO}(x)=\text{IWAE}_{1}(x)+\sum_{k=1}^{K}\frac{\Delta_{k}(x)}{\mathbb{P}(\mathcal{K}\geq k)}\text{, where }K\sim p_{k}(K)\text{ and }K\in\mathbb{N}^{+}\text{.}
\end{equation}
There are $\mathbb{P}(\mathcal{K}=K)=p_{k}(K)$ and $\Delta_{k}(x)=\text{IWAE}_{k+1}(x)-\text{IWAE}_{k}(x)$, where $\text{IWAE}_{k}(x)$ is the importance-weighted auto-encoder~\citep{burda2015importance}, defined as
\begin{equation}
    \text{IWAE}_{k}(x)=\log\frac{1}{k}\sum_{j=1}^{k}\frac{p_{\theta}(x|z_{j})p_{\theta}(z_{k})}{q_{\phi}(z_{k}|x)}\text{, where }z_{k}\stackrel{i.i.d}{\sim}q_{\phi}(z|x)\text{.}
\end{equation}
Now we can derive an upper bound for our FD reformulated objectives exploiting SUMO. To see how to achieve this, we can first derive a tractable lower bound for the first-order squared term as
\begin{equation}
    \begin{split}
    &\mathbb{E}_{p_{\epsilon}(v)}\mathbb{E}_{p_{\textup{data}}(x)}\left[\left(\log p_{\theta}(x+v)-\log p_{\theta}(x-v)\right)^{2}\right]\\
  =&\mathbb{E}_{p_{\epsilon}(v)}\mathbb{E}_{p_{\textup{data}}(x)}\left[\left(\mathbb{E}_{p_{k}(K_{1}),p_{k}(K_{2})}\left[\text{SUMO}(x+v;K_{1})-\text{SUMO}(x-v;K_{2})\right]\right)^{2}\right]\\
       \leq& \mathbb{E}_{p_{\textup{data}}(x)}\mathbb{E}_{p_{k}(K_{1}),p_{k}(K_{2})}\mathbb{E}_{p_{\epsilon}(v)}\left[\left(\text{SUMO}(x+v;K_{1})-\text{SUMO}(x;K_{2})+2\right)^{2}\right]\text{,}
    \end{split}
    \label{app:first}
\end{equation}
as well as a tractable unbiased estimator for the second-order term as
\begin{equation}
    \begin{split}
    &\mathbb{E}_{p_{\epsilon}(v)}\mathbb{E}_{p_{\textup{data}}(x)}\left[\log p_{\theta}(x+v)+\log p_{\theta}(x-v)-2\log p_{\theta}(x)\right]\\
  =&\mathbb{E}_{p_{\textup{data}}(x)}\mathbb{E}_{p_{k}(K_{1}),p_{k}(K_{2}),p_{k}(K_{3})}\mathbb{E}_{p_{\epsilon}(v)}\left[\text{SUMO}(x+v;K_{1})+\text{SUMO}(x-v;K_{2})\right.\\
  &\left.-2\cdot\text{SUMO}(x;K_{3})\right]\text{,}
    \end{split}
    \label{app:second}
\end{equation}
where we adjust the order of expectations to indicate the operation sequence in implementation. According to Eq.~(\ref{app:first}) and Eq.~(\ref{app:second}), we can construct upper bounds for our FD-SSM and FD-DSM objectives, and then train the LVMs via minimizing the induced upper bounds. In comparison, when we directly estimate the gradient-based terms $v^{\top}\nabla_{x}\log p_{\theta}(x)$ and $v^{\top}\nabla_{x}^{2}\log p_{\theta}(x)v$, we need to take derivatives on the SUMO estimator, which requires technical derivations~\citep{Luo2020SUMO}.

\subsection{Connection to MPF}
We can provide a naive FD reformulation for the SSM objective as
\begin{equation}
\begin{split}
    \!\!\!\mathcal{R}(\theta)&\!=\!\frac{1}{2\epsilon^2}\mathbb{E}_{p_{\textup{data}}(x)}\mathbb{E}_{p_{\epsilon}(v)}\!\left[\left(\log p_{\theta}(x\!+\!v)\!-\!\log p_{\theta}(x)\right)^{2}\!+\!4(\log p_{\theta}(x\!+\!v)\!-\!\log p_{\theta}(x))\right]\\
    &\!=\!\frac{1}{\epsilon^2}\mathbb{E}_{p_{\textup{data}}(x)}\mathbb{E}_{p_{\epsilon}(v)}\!\left[\frac{1}{2}\left(v^{\top}\nabla_{x}\log p_{\theta}(x)\right)^{2}\!+\!v^{\top}\nabla^{2}_{x}\log p_{\theta}(x)v\!+\!o(\epsilon^2)\right]\!=\!\mathcal{J}_{\text{SSM}}(\theta)\!+\!o(1)\text{.}\!\!
    \label{reformulation}
\end{split}
\end{equation}
Minimum probability flow (MPF)~\citep{sohl2011new} can fit probabilistic model parameters via establishing a deterministic dynamics. For a continues state space, the MPF objective is
\begin{equation*}
    \text{K}_{\text{MPF}}=\mathbb{E}_{p_{\textup{data}}(x)}\int g(y,x)\exp\left(\frac{E_{\theta}(x)-E_{\theta}(y)}{2}\right)dy\text{,}
\end{equation*}
where $E_{\theta}(x)=-\log p_{\theta}(x)-\log Z_{\theta}$ is the energy function. Let $B_{\epsilon}(x)=\{x+v|\|v\|_{2}\leq \epsilon\}$ and we choose $g(y,x)=\mathbbm{1}(y\in B_{\epsilon}(x))$ be the indicator function, then the MPF objective becomes
\begin{equation*}
    \hat{\text{K}}_{\text{MPF}}=V_{\epsilon}\mathbb{E}_{p_{\textup{data}}(x)}\mathbb{E}_{p_{\epsilon}(v)}\left[\exp\left(\frac{\log p_{\theta}(x+v)-\log p_{\theta}(x)}{2}\right)\right]\text{,}
\end{equation*}
where $V_{\epsilon}$ denotes the volume of $d$-dimensional hypersphere of radius $\epsilon$. Let $\Delta_{\theta}(x,v)=\log p_{\theta}(x+v)-\log p_{\theta}(x)$, then we can expand the exponential function around zero as
\begin{equation*}
    \hat{\text{K}}_{\text{MPF}}\!=\!V_{\epsilon}\mathbb{E}_{p_{\textup{data}}(x)}\mathbb{E}_{p_{\epsilon}(v)}\!\left[1\!+\!\frac{\Delta_{\theta}(x,v)}{2}\!+\!\frac{\Delta_{\theta}(x,v)^2}{8}\!+\!o(\Delta_{\theta}(x,v)^2)\right]\!=\!\frac{V_{\epsilon}\epsilon^2}{4}\left[\mathcal{R}(\theta)\!+\!o(1)\right]+V_{\epsilon}\text{,}
\end{equation*}
where the second equation holds because $\Delta_{\theta}(x,v)=\Theta(\epsilon)$. In this case, after removing the offset and scaling factor, the objective of MPF is directly equivalent to $\mathcal{R}(\theta)$ as to an $o(1)$ difference.

\section{Implementation details}
In this section, we provide a pseudo code for the implementation of FD formulation for both SSM and DSM. Then we provide the specific details in our experiments.

\subsection{Pseudo codes}
The pseudo code of FD-SSM is as follows:
\begin{lstlisting}[language=python,frame=shadowbox]
cat_input = concatenate([data, data + v, data - v], dim=0)
energy_output = energy(cat_input)
energy1, energy2, energy3 = split(energy_output, 3, dim=0)

loss1 = (energy2 - energy3)**2 / 4 
loss2 = (-energy2 - energy3 + 2 * energy1) * 2
FD_SSM_loss = (loss1 + loss2).mean() / eps ** 2
\end{lstlisting}

The pseudo code of FD-DSM is as follows:
\begin{lstlisting}[language=python,frame=shadowbox]
pdata = data + noise
cat_input = concatenate([pdata + v, pdata - v], dim=0)
energy_output = energy(cat_input)
energy1, energy2 = split(energy_output, 2, dim=0)

loss1 = (energy2 - energy1) * 0.5
loss2 = sum(v * noise/sigma, dim=-1)
FD_DSM_loss = ((loss1 + loss2)**2).mean() / eps ** 2
\end{lstlisting}

\subsection{Implementation details and definitions}

\textbf{DKEF} defines an unnormalized probability in the form of $\log \tilde{p}(x) = f(x) + \log p_0(x)$, with $p_0$ is the base measure. $f(x)$ is defined as a kernel function $f(x) = \sum_{i=1}^{N}\sum_{j=1}^{N_j}k_i(x, z_j)$, where $N$ is the number of kernels, $k(\cdot, \cdot)$ is the kernel function, and ${z_j}_{0<j<N_j+1}$ are $N_j$ inducing points.
We follow the officially released code from~\citet{song2019sliced}. Specifically, we adopt three Gaussian RBF kernel with the extracted by a three-layer fully connected neural network~(NN) with 30 hidden units. The width parameters for the Gaussian kernel is jointly optimized with the parameters of the NN. We apply the standard whitening process during training following~\citet{wenliang2019learning} and~\citet{song2019sliced}. We adopt Adam optimizer~\cite{kingma2014adam} with default momentum parameters and the learning rate is $0.01$. The only extra hyper-parameter $\epsilon$ in the finite-difference formulation is set to $0.1$.

\textbf{Deep EBM} directly defines the energy function with unnormalized models using a feed forward NN $f(\cdot)$ and the probability is defined as $p(x) = \frac{\exp(-f(x))}{\int \exp(-f(x)) dx}$.
The learning rate for DSM is $5\times 10^{-5}$ and the learning rate for SSM is $1\times 10^{-5}$ since the variance of SSM is larger than DSM. The optimizer is Adam with $\beta_1=0.9$ and $\beta_2=0.95$. The sampling method is annealed SGLD with a total of $2,700$ steps. The $\epsilon$ in the finite-difference formulation is set to $0.05$. When training with annealed DSM, the noise level is an arithmetic sequence from $0.05$ to $1.2$ with the same number of steps as the batch size. The default batch size is $128$ in all our experiments unless specified. The backbone we use is an 18-layer ResNet~\cite{he2016deep} following~\citet{li2019annealed}. No normalizing layer is used in the backbone and the output layer is of a generalized quadratic form. The activation function is ELU. All experiments adopt the ResNet with 128 filters. During testing, we randomly sample $1500$ test data to evaluate the exact score matching loss.

\textbf{NICE} is a flow-based model, which converts a simple distribution $p_0$ to the data space $p$ using a invertible mapping $f$. In this case, the probability is defined as $\log p(x) = \log p_0(z) + \log \det(\frac{\partial z}{\partial x})$, where $z=f^{-1}(x)$ and $\det(\cdot)$ denotes the determinant of a matrix.
The NICE model has 4 blocks with 5 fully connected layers in each block. Each layer has $1,000$ units. The activation is Softplus. Models are trained using Adam with a learning rate of $1\times 10^{-4}$. The data is dequantized by adding a uniform noise in the range of $[-\frac{1}{512}, \frac{1}{512}]$, which is a widely adopted dequantization method for training flow models. The $\epsilon$ in the finite-difference formulation is set to $0.1$.

\textbf{NCSN} models a probability density by estimating its score function, i.e., $\nabla_x \log p(x)$, which is modeled by a score net. We follow~\citet{song2019generative} and provide an excerpt on the description of the model architecture design in the original paper: "We use a 4-cascaded RefineNet~\citep{lin2017refinenet} and pre-activation residual blocks. We replace the batch normalizations with CondInstanceNorm++~\citep{dumoulin2016learned}, and replace the max-pooling layers in Refine blocks with average pooling. Besides, we also add CondInstanceNorm++ before each convolution and average pooling in the Refine blocks. All activation functions are chosen to be ELU. We use dilated convolutions~\citep{yu2017dilated} to replace the subsampling layers in residual blocks, except the first one. Following the common practice, we increase the dilation by a factor of 2 when proceeding to the next cascade. For CelebA and CIFAR-10 experiments, the number of filters for layers corresponding to the first cascade is 128, while the number of filters for other cascades are doubled. For MNIST experiments, the number of filters is halved."

% \begin{table}[t]
%   \centering
%   \vspace{-0.6cm}
%   \caption{The results of training implicit encoders for VAE and WAE on the MNIST and CelebA datasets. The models are trained for 100K iterations with the batch size of 128.}
%   \vspace{0.2cm}
%     \begin{tabular}{c|c|cc|cc}
%     \toprule
%          \multirow{2}{*}{Model} & \multirow{2}{*}{Algorithm}& \multicolumn{2}{c|}{MNIST} & \multicolumn{2}{c}{CelebA} \\
% %\cline{3-6}
%          &  & NLL &  Time & FID &  Time \\
%     \midrule
%     \multirow{2}{*}{VAE}& SSMVR & 89.58 & 5.04 ms & 62.76 & 14.9 ms \\
%     &\textbf{FD-SSMVR} & 88.96 & \textbf{3.98 ms} & 64.85 & \textbf{9.38 ms} \\
%     \midrule
%     \multirow{2}{*}{WAE} & SSMVR & 90.45 &   0.55 ms    &   54.28    &  1.30 ms\\
%     & \textbf{FD-SSMVR} & 90.66 &   \textbf{0.39 ms}    &     54.67  &  \textbf{0.81 ms}\\
%     \bottomrule
%     \end{tabular}
%   \label{tab:vae_wae}%
%   \vspace{-0.3cm}
% \end{table}%

\subsection{Details of the results on out-of-distribution detection}
For out-of-distribution (OOD) detection, we apply the typicality~\citep{nalisnick2019detecting} as the detection metric. Specifically, we first use the training set $\mathcal{D}_{\textup{train}}$ to approximate the entropy of model distribution as
\begin{equation}
    \mathbb{H}[p_{\theta}(x)]\approx \frac{1}{N}\sum_{x\in\mathcal{D}_{\textup{train}}}-\log p_{\theta}(x)\textup{,}
\end{equation}
where $|\mathcal{D}_{\textup{train}}|=N$ indicates the number of elements in the training set. Then give a set of test data $\mathcal{D}_{\textup{test}}$, where we control $|\mathcal{D}_{\textup{test}}|=M$ as a hyperparameter, then we can calculate the typicality as
\begin{equation}
    \left|\left(\frac{1}{M}\sum_{x\in\mathcal{D}_{\textup{test}}}-\log p_{\theta}(x)\right)-\mathbb{H}[p_{\theta}(x)]\right|\text{.}
    \label{typ}
\end{equation}
Note that the metric in Eq.~(\ref{typ}) naturally adapt to unnormalized models like EBMs, since there is
\begin{equation}
    \begin{split}
        &\frac{1}{M}\sum_{x\in\mathcal{D}_{\textup{test}}}-\log p_{\theta}(x)=Z_{\theta}+\frac{1}{M}\sum_{x\in\mathcal{D}_{\textup{test}}}-\log \widetilde{p}_{\theta}(x)\text{;}\\
        &\frac{1}{N}\sum_{x\in\mathcal{D}_{\textup{train}}}-\log p_{\theta}(x)=Z_{\theta}+\frac{1}{N}\sum_{x\in\mathcal{D}_{\textup{train}}}-\log \widetilde{p}_{\theta}(x)\text{,}
    \end{split}
\end{equation}
where the intractable partition function $Z_{\theta}$ can be eliminated after subtraction in Eq.~(\ref{typ}). Thus we can calculate the typicality for EBMs as
\begin{equation}
    \left|\left(\frac{1}{M}\sum_{x\in\mathcal{D}_{\textup{test}}}-\log \widetilde{p}_{\theta}(x)\right)-\left(\frac{1}{N}\sum_{x\in\mathcal{D}_{\textup{train}}}-\log \widetilde{p}_{\theta}(x)\right)\right|\text{.}
\end{equation}
As shown in \citet{nalisnick2019detecting}, a higher value of $M$ usually lead to better detection performance due to more accurate statistic. Thus to have distinguishable quantitative results, we set $M=2$ in our experiments. As to training the deep EBMs for the OOD detection, the settings we used on SVHN and CIFAR-10 are identical to those that we introduced above. On the ImageNet dataset, the images are cropped into a size of 128$\times$128, and we change the number of filters to 64 limited by the GPU memory. On SVHN and CIFAR-10, the models are trained on two GPUs, while the model is trained on eight GPUs on ImageNet. For all datasets, we use $N=50,000$ to estimate the data entropy and randomly sample $1000M$ test samples to conduct OOD detection.

\vspace{-0.1cm}
\subsection{Results on the VAE / WAE with implicit encoders}
\vspace{-0.1cm}
VAE / WAE with implicit encoders enable more flexible inference models. The gradient of the intractable entropy term $H(q)$ in the ELBO can be estimated by a score net. We adopt the identical neural architectures as in~\citet{song2019sliced}. The encoder, decoder, and score net are both 3-layer MLPs with 256 hidden units on MNIST and 4-layer CNNs on CelebA. For MNIST, the optimizer is RMSProp with the learning rate as $1\times 10^{-3}$ in all methods. The learning rate is $1\times 10^{-4}$ on CelebA. All methods are trained for 10K iterations. The $\epsilon$ in the finite-difference formulation is set to $0.1$.

\end{document}